\documentclass[3p]{elsarticle}

\usepackage{lineno,hyperref}

\usepackage[labelfont=bf]{caption}
\usepackage{caption}
\DeclareCaptionLabelSeparator{point}{.~}
\usepackage{amsmath}
\usepackage{algorithm}
\usepackage[noend]{algpseudocode}
\usepackage{booktabs}
\usepackage[skip=0.1\baselineskip]{caption}
\usepackage{subcaption}
\usepackage{multirow}
\usepackage{tabu}
\usepackage{xcolor}
\usepackage{graphicx}

\journal{arXiv.}

\begin{document}
\begin{frontmatter}

\title{An Explainable Artificial Intelligence Approach for Unsupervised Fault Detection and Diagnosis in Rotating Machinery}

\author[mymainaddress]{Lucas C. Brito\corref{mycorrespondingauthor}}
\cortext[mycorrespondingauthor]{Corresponding author}
\ead{lucas.brito@ufu.br | brito.lcb@gmail.com}
\author[mysecondaryaddress]{Gian Antonio Susto}
\author[mythirdaryaddress]{Jorge N. Brito}
\author[mymainaddress]{Marcus A.V. Duarte}

\address[mymainaddress]{School of Mechanical Engineering, Federal University of Uberlândia, Av. João N. Ávila, 2121, Uberlândia, Brazil}
\address[mysecondaryaddress]{Department of Information Engineering, University of Padova, Via Gradenigo 6/B, 35131, Padova, Italy}
\address[mythirdaryaddress]{Department of Mechanical Engineering, Federal University of São João del Rei, P.Orlando, 170, São João del Rei, Brazil}

\begin{abstract}
The monitoring of rotating machinery is an essential task in today's production processes. Currently, several machine learning and deep learning-based modules have achieved excellent results in fault detection and diagnosis. Nevertheless, to further increase user adoption and diffusion of such technologies, users and human experts must be provided with explanations and insights by the modules. Another issue is related, in most cases, with the unavailability of labeled historical data that makes the use of supervised models unfeasible. Therefore, a new approach for fault detection and diagnosis in rotating machinery is here proposed. The methodology consists of three parts: feature extraction, fault detection and fault diagnosis. In the first part, the vibration features in the time and frequency domains are extracted. Secondly, in the fault detection, the presence of fault is verified in an unsupervised manner based on anomaly detection algorithms. The modularity of the methodology allows different algorithms to be implemented. Finally, in fault diagnosis, Shapley Additive Explanations (SHAP), a technique to interpret black-box models, is used. Through the feature importance ranking obtained by the model explainability, the fault diagnosis is performed. Two tools for diagnosis are proposed, namely: unsupervised classification and root cause analysis. The effectiveness of the proposed approach is shown on three datasets containing different mechanical faults in rotating machinery. The study also presents a comparison between models used in machine learning explainability: SHAP and Local Depth-based Feature Importance for the Isolation Forest (Local-DIFFI). Lastly, an analysis of several state-of-art anomaly detection algorithms in rotating machinery is included.
\newline
\end{abstract}

\begin{keyword}
\texttt Anomaly Detection \sep Explainable Artificial Intelligence \sep Fault Detection \sep Fault Diagnosis \sep Rotating Machinery \sep Condition Monitoring 

\end{keyword}
\end{frontmatter}


\textit{This work has been submitted  for possible publication. Copyright may be transferred without notice, after which this version may no longer be accessible.}

\section{Introduction and Related Work}

The study of artificial intelligence (AI) techniques applied in the monitoring of rotating machinery is a topic in continuous development and of great interest by both researchers and industrial engineers. More and more, industries are adopting more sophisticated technologies for monitoring to increase the reliability and availability of the machines, and, consequently, remaining competitive in the globalized economy.

As detailed by \cite{LIU201833} there are three basic tasks of fault diagnosis: (1) determining whether the equipment is normal or not; (2) finding the incipient fault and its reason; (3) predicting the trend of fault development. It is clear that when determining the type of fault and its reason (task 2), consequently the answer on the condition of the equipment is obtained (task 1). However, the AI models usually used to classify the type of fault require to be trained with labeled data (supervised training), and examples for all conditions, which in most of the cases is not available in the industry \cite{carletti2019explainable}. In addition, motivated by the recent advances in Deep Learning (DL), the vast majority of AI technologies lack of explainability traits and they require a large volume of data labeled for both normal and fault conditions, dramatically limiting their industry application. 

Although the field of rotating machinery monitoring is widely developed, a small number of approaches have been presented based on unsupervised anomaly detection, in relation to the vast majority focused on classification and prognostics, as shown in the review works \cite{LIU201833, Kumar,LEI2020106587,STETCO2019620}. A detailed review of AI for fault detection in rotating machines is presented in \cite{LIU201833}: most of the 100 references cited there refer mainly to the fault classification. In their review, the main models present in literature were: Artificial Neural Networks (ANNs), k-Nearest Neighbor (kNN), Naives Bayes, Support Vector Machines (SVM) and DL-based approaches. In \cite{Kumar} a review of the main machine learning (ML) and DL techniques applied in the monitoring of induction motors is presented, aiming to detect faults such as: broken bars, bearings, stator faults and eccentricity. Among more than 100 references cited, the vast majority refers to classification and the main models used are: ANNs, Decision Trees, k-NN, SVM and DL-based approaches.

More recently, a broad review with more than 400 citations, focused on AI applications for fault detection is presented \cite{LEI2020106587}. The authors provide a historical overview, in addition to current developments and future prospects. Among the revised ML methods employed in the field, the authors recognized the following as the most commonly adopted: ANNs, Decision Trees, kNN, Probabilistic Graphical Model (PGM) and SVM. Moreover, the following DL approaches are taken into consideration: Autoencoders (AEs), Convolutional Neural Networks (CNN), Deep Belief Network (DBN), Residual Neural Networks (ResNet). As the other cited review works, the references in \cite{LEI2020106587} mainly focused on classification of the type of fault. The authors also confirm the dependence on real and labeled data from the machine under analysis. In addition to highlighting the recent and future importance in the Intelligent Fault Diagnosis (IFD) scenario of explainable models, with increasing interest starting from 2017. Finally, they mention that traditional ML models should not be abandoned despite the recent advances of DL: this is because it is still worth investigating statistical learning in IFD with the big data revolution, since the theories of statistical learning have rigorous theoretical bases, which promote the construction of diagnostic models with parameters, characteristics and results that are easy to understand.

Anomaly detection is the process of identifying unexpected events in the dataset, which are different from normal. In general, the signals generated by a fault have characteristic patterns that are different from normal and indicate a change in the behavior of the machine. Using a method that indicates changes in the current condition of the equipment, does not need a labeled historical dataset for training and provides explainability of the results can be the solution to the mass dissemination of artificial intelligence methods in the industrial environment for monitoring rotating machinery.

Among the references studied, there are very few studies involving anomaly detection with unsupervised approaches in the monitoring of rotating machinery. Authors in \cite{ogata} used Fourier local autocorrelation (FLAC) and Gaussian Mixture Model (GMM) approach (based on class cluster) to extract features in the time-frequency domain and to detect faults respectively; in the same work, vibration signals were used to detect faults in wind turbine components. The authors showed that the use of features extracted from FLAC improves the model's performance, making it possible to detect anomalies in even more complicated cases, such as low speeds, where conventional features do not present such satisfactory results. In \cite{vonBirgelen} an application of the Self Organizing Maps (SOM) for anomaly detection was presented, showing its effectiveness in detecting variations when the component fails: uses case related to cyber-physical system components (bearings and blades) were exploited. In \cite{Amruthnath2} the authors proposed the use of different methods of ML using vibration signals from a fan for unsupervised detection of incipient fault. The algorithms used were: PCA T2 statistic, Hierarchical clustering, K-Means, Fuzzy C-Means clustering. Finally, they presented a comparison of the models, showing the feasibility of implementing them in the monitoring of machines. Other studies \cite{zhangmssp,Hasegawa1,Hasegawa2} propose anomaly detection approaches in the monitoring of rotating machinery based on combinations of different techniques with variations of GMM. To the best of our knowledge, the vast majority of state-of-the-art ML unsupervised anomaly detection, have never being used, e.g.,  Isolation Forest (IF), Local Outlier Factor (LOF), Angle-Based Outlier Detection (ABOD) etc. 

Another important aspect in the field that has not been fully explored yet is the one related to interpretability of ML-based monitoring solutions in equipment machinery: as argued above, without providing explainable results to the user, even when ML-based modules provide excellent results in historical data, AI  models are unlikely to be applied in real-world scenarios \cite{molnar2020interpretable}. Moreover, as mentioned by \cite{LEI2020106587}, collecting labeled data from machines generates a high cost, and consequently unlabeled data is the majority in engineering scenarios. Therefore, using an AD (anomaly detection) model that works with unlabeled data and provides explainability is essential to enable large-scale implementation of AI in the monitoring of rotating machinery.

Recently studies are being developed with a focus on Explainable Artificial Intelligence (XAI). In order to explain black-box models, different methods can be used according to the ML model in use \cite{du2019techniques}. In general, the methods provide information to understand how the model performs fault detection, which can be, for example, a ranking of the most important features, the model weight relevance or the most significant points in the underlying signals \cite{doshi2017towards}. Despite the current interest, the vast majority of studies are focused on explainability for DL models and mostly on fault classification. More information can be found in the articles available on the topic \cite{xaidl1,xaidl2,xaidl3,xaidl4,xaidl5,xaidl7,xaidl8,xaidl9,xaidl10,Saeki}. Among the references researched, only \cite{Saeki, Hendrickx} address the explainability of the model in anomaly detection, being \cite{Saeki} based on DL. \cite{Hendrickx} presented a methodology for detecting anomalies in electric motors (voltage unbalance) using a set of similar equipment through electrical and vibration signature. The authors use generic building blocks and present advantages of not needing historical data, incorporating human knowledge. Despite the interesting approach, it is noted that for its use it is necessary to have data from more than one machine, so that they can be compared, making applications on single machines unfeasible. 

In this paper, a new approach for fault detection and diagnosis in rotating machinery is proposed. In the first part, the vibration features in the time and frequency domains are extracted. Secondly, in the fault detection, the presence of fault is verified in an unsupervised manner based on anomaly detection algorithms. Finally, in fault diagnosis, Shapley Additive Explanations (SHAP), a technique to interpret black-box models, is used. Through the feature importance ranking obtained by the model's explainability, the fault diagnosis is performed. Two approaches of diagnosis are proposed, namely: unsupervised classification and root cause analysis.

The main contributions of the proposed approach are: i) unsupervised identification of the fault in rotating machinery through vibration analysis; ii) unsupervised classification of the type of fault in rotating machinery, based on the analysis of the features relevance; iii) possibility of performing root cause analysis when the features may be related to more than one fault and the unsupervised classification is not feasible; iv) a new contribution to the study of XAI and novel application in fault diagnosis for rotating machinery is presented based on SHAP and Local-DIFFI; v) possibility to be applied in different types of faults; vi) possibility to change models according to the dataset; vii) industrial applications.

To the best of the authors' knowledge, this is the first study to compare and analyze unsupervised state-of-the-art anomaly detection algorithms for monitoring rotating machinery. In addition to providing explainability about the ML models used and proposing a new approach to perform unsupervised classification or root causes analysis.

The remainder of this paper starts with a brief explanation about the machine learning and XAI methods used in Section 2. The proposed approach is presented in Section 3. Experimental procedure is shown in Section 4. Analysis of the experimental results are given in Section 5. Finally, Section 6 concludes this paper.

\section{Methodologies}

\subsection{Anomaly Detection Algorithms} \label{sec:ad}

In this sub-Section we will provide a brief overview on the data-driven unsupervised Anomaly Detection (AD) algorithms compared in this work. 

Anomaly detection (also known as outlier detection\footnote{The terms 'Anomaly' and 'Outlier' will be treated in the same way in this work.}) refers to the task of identifying rare observations which differ from the general ('normal') distribution of a data at hand \cite{zhao2019pyod}. Anomaly Detection approaches have the capability of summarizing the status of a multivariate systems with a unique quantitative indicator, that is typically called \emph{Anomaly Score} (AS)\footnote{Other authors refer to the concept of Anomaly Score with various names like for example Health Factor or Deviance Index.}: while many approaches provide guidelines on how to define outliers based on the AS, the quantitative nature of the AS indeces allowed to implement different strategies that allow to govern the trade-off between false positives and false negatives depending on the application at hand.  While no applications to the best of our knowledge have been presented in the field of rotating machinery monitoring using vibration data and state-of-art models (that will be introduced in the rest of the Section), anomaly detection approaches have been successfully applied in various areas like biomedical engineering \cite{meneghetti2018data}, fraud detection \cite{rai2020fraud}, oil and gas \cite{barbariol2020self}.

Algorithms are arranged by increasing year of presentation.

\subsubsection{k-Nearest Neighbors (kNN)}

k-nearest neighbor (kNN) is a simple and popular method used for supervised tasks of classification and regression. In the context of AD, kNN can be also employed: given a sample, the distance to its kth-nearest neighbor can be considered as AS \cite{knn_1}. More formally, the anomaly score \cite{knn_2} is then defined as:
\begin{flalign}
   & s_\text{kNN}(x) =  D^k(x) &
\end{flalign}
where \(D^k(x)\) denotes the distance of the \(k^{th}\) nearest neighbor from observation \(x\). The distance function can be any metric distance function. The most common methods for selecting distance function are: largest distance, where the distance to the \(k^{th}\) neighbor is used as the AS; mean distance, where the AS is the average of all \(k\) neighbors; median distance, which uses the median of the distance to \(k\) neighbors as AS.

\subsubsection{Minimum Covariance Determinant (MCD)}

The minimum covariance determinant (MCD) is a robust estimator of multivariate locations and its goal is to find \(n\) instances (out of \(N\)) whose covariance matrix has the lowest determinant \cite{mcd_1}. In the context of AD, MCD is used with Mahalanobis distance (MD), a well-known distance metric of a point from a distribution: first a minimum covariance determinant model is fitted and then the Mahalanobis distance is used as AS. Since the parameters required by MD are unknown (mean and covariance matrix), the MCD model is used to estimate them, and then the MD can be calculated as follows: 
\begin{flalign}
    & s_\text{MCD}(x) = d(x,\bar{x},Cov(X)) = \sqrt{(x-\bar{x})'Cov(X)^{-1}(x-\bar{x})} &
\end{flalign}
where \(\bar{x}\) is the sample mean and \(Cov (X)\) is the sample covariance matrix. If data are assumed centered not normalized, the robust location and covariance are directly computed with the FastMCD algorithm without additional treatment. Otherwise, the support of the robust location and the covariance estimate are computed, and a covariance estimate is recomputed from it, without centering the data \cite{zhao2019pyod}.

\subsubsection{Local Outlier Factor (LOF) and Cluster-based Local Outlier Factor (CBLOF)}

LOF \cite{lof} is a density-based approach for AD; such class of approaches are based on the study of local neighborhoods of the data points under exam: an observation is a dense region is considered as a normal data point (also referred in the literature as an \emph{inlier}), while observations in low-density regions are anomalies.

The LOF procedure involves two steps: (i) evaluating the so-called Local Reachability Density; (ii) evaluating the AS $s_\text{LOF}$. the Local Reachability Density of a data point $x$ in its $k$-neighborhood $\mathcal{N}_k(x)$ (the space where the $k$ other data points closest to $x$ are living) is defined as:
\begin{flalign}
   & \text{LRD}_k(x) = \frac{k}{\sum_{k}(y \in \mathcal{N}_k(x)) r_k(x, y)} &
\end{flalign}
where $r_x(x, y) = \text{max}\{d_k(x), d(x,y)\}$ is the so-called reachability distance and $d_k(x)$ is the distance from $x$ of its $k$-th nearest neighbor. The reachability distance just defined is used instead of the distance $d(x,y)$ in order to reduce statistical fluctuations/noise in the evaluation of the AS $s_\text{LOF}$; the AS is in fact defined as:
\begin{flalign}
    & s_\text{LOF}(x) = \frac{1}{k} \sum_{y \in \mathcal{N}_k(x)} \frac{\text{LRD}_k(y)}{\text{LRD}_k(x)} &
\end{flalign}
The above defined anomaly score can assume values between 0 and $\infty$, however, a value around 1 (or lower than 1) indicates that the data point $x$ is somehow similar to its neighbors and it can be therefore considered as an inlier; a value of  $s_\text{LOF}$ larger than 1 indicates instead a case in which the data point under exam can be considered as an outlier. For more details we refer the interest readers to \cite{lof}.

LOF is a classic approach to AD and extended versions of the algorithms have been proposed over the years \cite{kriegel2009loop, schubert2012evaluation}: in this work we consider the popular Cluster-based LOF (CBLOF). The CBLOF \cite{cblof} algorithm\footnote{In the original paper \cite{cblof}, the authors indicated with 'CBLOF' the AS computed with the 'FindCBLOF' algorithm. Nevertheless the community has been referring also to the algorithm by the name 'CBLOF': in this work we will follow this naming convention for CBLOF and for other AD approaches.} for AD is an extended version of LOF that exploits a clustering procedure before applying the LOF algorithm: the underlying idea of this approach is to overcome a known problem in LOF that has some difficulties in dealing with data that are clustered.

First, a clustering algorithm (typically $k$-means) is used to partition the dataset into $k$ disjointed clusters $C = \{C_1, \ldots C_k\}$. Each data instance is assigned with an AS $s_{\text{CBLOF}}$ based on the size of the cluster it was assigned to: the method uses two cluster types that are called 'small cluster' ($SC$) and 'large cluster' ($LC$) based on the cardinality of the cluster. The coefficients for deciding small and large clusters are given by the numeric parameters \(\alpha\) and \(\beta\). Where $b$ is the boundary of a cluster, the anomaly score for a data point \(x\) is defined as:
\begin{flalign}
& s_{\text{CBLOF}}(x)= \begin{cases}
    |C_i|^{*}min(d(x,C_j)), \text{where} \; x \in C_i, C_i \in SC\; \text{and}\; C_j \in LC\; \text{for}\; j = 1\; to\; b \\
    |C_i|^{*}(d(x,C_i)), \text{where} \; x \in C_i\; \text{and}\; C_i \in LC
\end{cases}&
\end{flalign}

\subsubsection{One-class Support Vector Machines (OCSVM)}

One-Class Support Vector Machine \cite{ocsvm} is an extension for AD of the popular approach for classification known as Support Vector Machine. The training data is projected to a high-dimensional space and the hyperplane that best separates the points from the origin is determined. When evaluating a new sample, if it lays within the frontier-delimited subspace, it is considered to come from the same population and therefore it is considered as an inlier; otherwise, the data point is considered as an anomaly by the approach. 

As in SVM, kernel functions are used to produce non-linear hyperplanes; different kernels can be used: linear, polynomial, sigmoid, gaussian. In this work, the kernel coefficient for gaussian, polynomial and sigmoid will be called \(gamma\) and the parameter to define an upper bound on the fraction of training errors and a lower bound of the fraction of support vectors, \(nu\).

\subsubsection{Feature Bagging (FB)}

Feature Bagging is the combination of multiple outlier detection algorithms using different set of features \cite{fb}. Every outlier detection algorithm uses a small subset of features that are randomly selected from the original feature set. Any AD approach can be used as the base estimator. Using a cumulative sum approach, each AS generated by each outlier detector used is combined in order to find a final AS and described as:
\begin{flalign}
    & s_\text{final}(x) =  \sum_{t=1}^{T}s_t(x) &
\end{flalign}
where the final anomaly score \(s_\text{final}(x)\)is the sum of all anomaly scores \(s_t\) from all \(T\) iterations on each outlier detector used. The number of base estimators in the ensemble and the number of features to draw from X to train each base estimator can be adjusted. Moreover, the final combination of the AS can be performed by the averaging all models or taking the maximum scores.

\subsubsection{Angle-based Outlier Detector (ABOD) and Fast-ABOD}

Differently from the methods for detection outliers based on distances or distributions, Angle-based Outlier Detector (ABOD) \cite{abod} exploits considerations made on the angles obtained by considering the data point under exam as vertex and all the possible couples of points considering the other data present in the dataset. The underlying idea is that outliers will form angles with other data points that are typically acute, while inliers will form angles of different types: from small angles to straight ones. For this reason, what is monitored as AS for a generic datapoint $x$ is the variance of the angles formed by $x$ as a vertex. 

The computation of the all the angles formed by all the possible triples in the dataset is a time consuming operation: for this reason, several approximated versions of the ABOD algorithm have been proposed over the years. In this work we will employ the approximation presented in the original paper \cite{abod} that is called Fast-ABOD that consider only the angles formed by the data point under exam and its $k$ nearest neighbors; in the Fast-ABOD formulation the anomaly score is computed as:
\begin{flalign}
   & s_{\text{Fast-ABOD}}(\Vec{A}) =  VAR_{\Vec{B},\Vec{C}\in N_k\Vec{A}}\bigg(\dfrac{\big \langle \overline{AB},\overline{AC}\big \rangle}{\parallel{\overline{AB}}\parallel^2\;\parallel{\overline{AC}}\parallel^2}\bigg) &
\end{flalign}
where \(\big \langle.,.\big \rangle\) indicates the scalar product, \(\Vec{A}\),\(\Vec{B}\) and \(\Vec{C}\) are the considered data points and VAR is the variance over the angles between the difference vector of \(\Vec{A}\) to all pairs of points in \(N_k(\Vec{A})\) weighted by the distance of the points, and \(N_k(\Vec{A})\) is the set of $k$ nearest neighbors of $A$. It is important to highlight that although the Fast-ABOD presents a better computational cost, the quality of the approximation depends on the number \(k\)-nearest neighbors.

\subsubsection{Isolation Forest (IF)}

Isolation Forest (iForest or IF), \cite{if_1,if_2}, uses the concept of \emph{isolation} instead of measuring distance or density to detect anomalies. The IF exploits a space partitioning procedure: the main idea underlying the approach is that an outlier will require less iterations than an inlier to be isolated, i.e., to find through the partitioning procedure a region of the space where only such observation lies in.

The partitioning procedure used by the IF is achieved through the creation of iTrees, binary trees that are the result of a \emph{random} partitioning procedure obtained by splitting the data based on one of their features at each iteration of the algorithm. Following the above stated fundamental idea of  IF, it is expected that the path to reach a leaf node from the root of an iTree will be shorter for outliers than for inliers; the anomaly score will be related to this path length: the shorter the more anomalous the data point. We underline that this procedure is done randomly: to achieve fast computation the features and the splitting points are chosen randomly; the drawback of this approach is that a single tree can give an estimate of the path length that has high variance: thus, similar to the popular Random Forest (that we remark is a supervised approach), an ensemble of $T$ trees is constructed in order
to provide a low-variance estimation. More in detail, an iTree is built as follows.
\begin{enumerate}
    \item A subsample of data $S \in X$ is randomly selected.
    \item A feature $v \in \{1, \ldots, p\}$ is randomly selected: a node in the tree is created and at this node the value of $v$ is used;
    \item A random threshold $\bar{v}$ on $v$ is chosen within the domain of the variable;
    \item Two children nodes are generated: one associated to the points with values for variable $v$ below $\bar{v}$ and one for those with value above;
    \item The points from 2 to 4 of this procedure are repeated until either a data point is isolated or a threshold on the maximum tree length is reached.
\end{enumerate}
After the iTrees are constructed, the AS score for a data point $x$ is computed as follows:
\begin{flalign}
    & s_{\text{IF}}(x) = 2^{\frac{-E(h(x))}{c}} &
\end{flalign}
where \(h(x)\) is the length of the path for a data point from its leaf to the root, \(E(h(x))\) is the average of \(h(x)\) in iTrees collection of iTrees and \(c\) is an adjustment factor which is set to the average
path length of unsuccessful searches in a binary search tree
procedure. Using the AS just defined, if instances return $s_{IF}$ very close to 1, then they are tagged as anomalies; on the other hand, values much smaller than 0.5 are quite safe to classify as normal instances, and values close to 0.5 then the entire sample does not really have any distinct anomaly \cite{if_2}.

iForest works well in high dimensional problems which have a large number of irrelevant attributes, and in situations where training set does not contain any anomalies. Given its high performance and the possibility to parallelize its computation (thanks to its ensemble structure), IF is probably the most popular AD approach: for this reason, we will consider, as it will be detailed in Section \ref{diffi}, a dedicated approach for providing interpretable traits to IF.

\subsubsection{Histogram-based outlier score (HBOS)}

HBOS is an AD approach based on histograms that was introduced for providing fast computation of an AS w.r.t. previously proposed AD methods.

The HBOS algorithm can be summarized as follows: univariate histograms for each single feature are computed (in case of numerical data a set of $k$ bins of equal size are used for each histograms). The number of bins \(k\) is an hyper-parameter that needs to be tuned; histograms are normalized to $[0,1]$ for each single feature; frequency (relative amount) of samples in a bin is used as density estimation; AS for each instance $x$ is computed as a product of the inverse of the estimated density:
\begin{flalign}
    & s_{\text{HBOS}}(x)  =  \sum_{i=0}^{p}log\left({\frac{1}{hist_i(x)}}\right) &
\end{flalign}
where \(p\) is the number of features and \(hist_i(x)\) is the density estimation. With such definition of the AS, with HBOS the outliers correspond to high values of $s_{\text{HOBS}}(x)$, while inliers to low values.  In this algorithm, two parameters are still employed and need to be tuned, being \(\alpha\) and the tolerance (\(tol\)). \(\alpha\) is a regulation factor to avoid overfitting and \(tol\) adjusts the flexibility while dealing the samples falling outside the bins.

\subsubsection{Lightweight on-line detector of anomalies (LODA)}

Lightweight on-line detector of anomalies (LODA) is based on the concept of supervised learning that shows that a collection of weak classifiers can result in a strong classifier. LODA is comprised of a collection of \(k\) one-dimensional histograms with \(n\_bins\), each approximating the probability density of input data projected onto a single projection vector \cite{loda}. The average of the logarithm of probabilities estimated on individual projection vector is LODA output, \(f(x)\), defined as:
\begin{flalign}
    & f(x) =  -\dfrac{1}{k}\sum_{i=1}^{k}log\;{\hat{p_i}(x^{T}w_i}), &
\end{flalign}
where \(p(x^{T}w_i)\) is the joint probability of projections, in other words, \(\hat{p_i}\) is the probability estimated by the \(ith\) hisogram, \(w_i\) the corresponding projection vector and \(x\) the sample. LODA sparse random projections can also defined by the user, here called \(n\_random cuts\). Due to its simplicity LODA is particularly useful in domain where a large number of samples need to be processed in real-time or in domains subject to concept drift. It can also be applied where the detector needs to be updated on-line \cite{loda}.

\subsubsection{Ensemble}

The ensemble method combines different algorithms to obtain a single final result. Knowing that ML models are sensitive to the types of data, ensemble methods are commonly used to increase the efficiency and robustness of the final result. Being \(H_i\) the result of each \(i^{th}\) base model, the sum of the \(k\) selected ones is the final result (FR) of the ensemble method, and the final decision (FD) is obtained by a majority voting, both described as:
\begin{flalign}
  &  FD =  \begin{cases}
    1, \;\text{if} \; FR \;> \;k/2 \\
    0, \;\text{otherwise.}
\end{cases} \text{, where} \; FR =  \sum_{i=1}^{k}H_i &
\end{flalign}

Where in this case, 1 indicates that the sample is an anomaly and 0 that the sample is normal.

\subsection{Explainable Artificial Intelligence (XAI)}

In this subsection the XAI approaches adopted in this work are revised.

\subsubsection{Shapley Additive Explanations (SHAP)}

Shapley Additive Explanations, \cite{shap} is a state-of-art and model-agnostic (it can be applied to any algorithm)  for interpreting ML predictions, both in unsupervised and supervised tasks. 

Based on Shapley values from coalitional game theory, SHAP provides a feature importance ranking which can be used to explain the ML model to the individual data point level: in the context of anomaly detection, having an ordered list of features can be really helpful for domain expert to enable an effective troubleshooting. The feature importance ranking is the result of the contribution of each feature to the final prediction of the model.

Since the Shapley values are expensive to obtain, SHAP approximates them of a conditional expectation function of the original model. The detailed mathematical formulation of SHAP can be retrieved at \cite{shap}.

\subsubsection{Local Depth-based Feature Importance for the Isolation Forest (Local-DIFFI)}\label{diffi}

Given the increased interest and popularity of IF, we chose to consider in this work also a model-specific approach for providing, like in SHAP, a feature importance ranking.

Local Depth-based Feature Importance for the Isolation Forest is the first model-specific method for interpretability in IF \cite{carletti2020interpretable}. While IF is one of the most commonly adopted AD algorithms, its structure and prediction lack in interpretability. To overcome this problem the Local-DIFFI method proposes an effective and computationally inexpensive approach to define local feature importance (LFI) in IF, computed as:
\begin{flalign}
    & LFI = \dfrac{I_o}{C_o}, &
\end{flalign}
where \(C_o\) is the features counter for the single predicted outlier \(x_o\) and \(Io\) is updated by adding the quantity \cite{carletti2020interpretable} while iterating over all the trees in the forest:
\begin{flalign}
   & \Delta = \dfrac{1}{h_t(X_o)} - \dfrac{1}{h_{max}} &
\end{flalign}
The model is a post-hoc method, which, due to its operation, preserves the performance of an established and effective AD algorithm (IF). An interesting property of Local-DIFFI is that, while achieving comparable results w.r.t. SHAP, its computing time is orders of magnitudes smaller than SHAP. The method proposes to provide additional information about a trained instance of the IF model with the main objective of increasing the users' confidence in the result obtained. Besides the local feature importance provided by Local-DIFFI, the method can also be used to provide global feature importance, namely DIFFI.

\section{Proposed Approach}

\begin{figure*}[ht]
  \centering
  \includegraphics[scale=0.55]{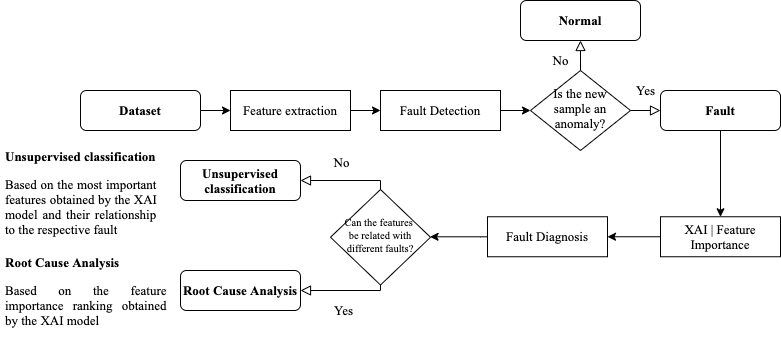}
  \caption{Framework of the proposed methodology.}
  \label{fig:framework}
\end{figure*}

The proposed methodology is depicted in Fig. \ref{fig:framework} and it is divided into three parts: 1) Feature extraction; 2) Fault detection: Anomaly Detection; 3) Fault diagnosis: Unsupervised classification / Root cause analysis. The vibration features are initially extracted based on the type of monitored component. The extracted features are divided into a training and testing group, and the hyperparameters of the anomaly detection models are tuned. The samples are evaluated in the fault detection part: if a fault (anomaly) is not detected, the analysis is completed; on the other hand, if the sample is a fault (anomaly), the most relevant features used to generate the result are evaluated through the model's explainability. In the fault diagnosis part, the features that indicate only the presence of fault, but do not indicate the type / location are disregarded (called general features, e.g., rms and kurtosis). For components that have unique fault specific features (e.g., bearing, gearbox), it is possible to perform an unsupervised classification based on the most relevant feature for the result. On the other hand, for analysis where the features may be related to more than one fault (e.g., misalignment and mechanical looseness), the most relevant features (feature ranking) for identifying the sample as an anomaly are presented, allowing the specialist to analyze the problem in more detail, namely root cause analysis. 
\subsection{Feature extraction}

One of the main reasons for the wide use of DL models in many tasks is that DL approaches implicitly implement a feature extraction procedure due to the DL architectures ability to learn discriminating features through non-linear relations performed within the model: avoiding the time consuming task of feature extraction is a captivating property for ML technologies developers. However, in the domain of rotating machinery, the vast majority of faults have already been studied and ad-hoc defined features that are informative for fault detection (that can be computed directly on the raw signals or after dedicated signal processing filters) have been developed by researchers over the years. For this reason, we have decided to base our approach on 'classic' ML techniques exploiting the wide knowlegde of filtering approaches and feature definitions provided by the literature.

Among the sensors used for monitoring rotating machinery, the vibration-based diagnostic method is the most popular and researched. The interest is justified by the fact that the vibration signals directly represent the dynamic behavior of the equipment \cite{feature01,feature02,feature03,feature04} and are a non-invasive technique. The features to detect faults in rotating machinery using vibration signals are commonly extracted from the time, frequency and time-frequency domains \cite{LEI2020106587}. 

(i) Among the most used in the time domain are: mean, standard deviation, rms (root mean square), peak value, peak-to-peak value. According to \cite{LEI20072280,LEI20101419}, these features can be affected by the speed and load of the machines, therefore, other features are also commonly used to fill this gap: shape indicator, skewness, kurtosis, crest factor, clearance indicator, etc., which are robust to the machine's operating conditions.

(ii) The features in the frequency domain are extracted from the frequency spectrum, for example: mean frequency, central frequency, energy in frequency bands, etc. Different information can be obtained that is not found or is hardly extracted in the time domain \cite{LEI2020106587}.

(iii) For the time-frequency domain, features such as entropy are usually extracted by Wavelet Transform (WT), Wavelet Packet Transform (WPT) and empirical model decomposition (EMD). These features are capable of reflecting the machine's health states in non-stationary operating conditions \cite{LEI2020106587}.

In this study, two approaches were combined in relation to the types of features. Firstly, general features were selected to indicate the presence of system fault and degradation. This approach does not allow the identification / location of the fault, but it allows to detect variations in the system in a global way, avoiding that a fault is not identified. Secondly, specific features commonly associated with the type of defect in the respective components were used to enable the identification / location of the fault. During the extraction of specific features, it must be defined whether the features are related to different faults or are unique, enabling the fault diagnosis through unsupervised classification or root cause analysis.

\subsection{Fault detection: Anomaly detection}

Identifying the fault is extremely important for production processes, and even more important when performed in an unsupervised manner, that is, without the presence of labeled data related to the fault modes in training set. In this part, the extracted features are divided into a training and testing group, and the hyperparameters of each AD model are adjusted. The samples are evaluated in unsupervised manner and identified as normal or fault (anomaly). If an anomaly is not considered, the analysis is completed. On the other hand, if the sample is an anomaly, the root cause analysis or fault classification based on the model's explainability is performed.

Different models used in the field of AD were studied. As is common knowledge, the performance of AD models are strongly related to the type of data available. While we report in the following the best approaches in our studies, the approach is generic and the user can modify the AD model in use if the expected performance is not achieved, without affecting its structure.

The different AD algorithms evaluated were the ones reported in Section \ref{sec:ad}: Clustering Based Local Outlier Factor (CBLOF), Local Outlier Factor (LOF), Isolation Forest (IF), Lightweight on-line detector of anomalies (LODA), Histogram-based Outlier Detection (HBOS), k-Nearest Neighbors (kNN), Fast - Angle-based Outlier Detector (FastABOD), Outlier Detection with Minimum Covariance Determinant (MCD), One-Class Support Vector Machine (OCSVM), Feature Bagging (FB) and Ensemble (combination of all models) available in \cite{zhao2019pyod}.

\subsection{Fault diagnosis: Unsupervised Classification / Root Cause Analysis}

Despite the advances in ML applications for fault diagnosis in rotating machines, the vast majority of methods are performed in a supervised manner. In other words, the methods use labeled data in the training to ensure that the model is able to distinguish between different classes of faults. The proposed methodology presents an approach where no training labels are necessary. The fault diagnosis is performed in an unsupervised manner based on the importance ranking obtained by the model explainability. Two different analysis are possible depending on the type of component being monitored, namely: unsupervised classification and root cause analysis. For faults that have unique characteristic features (e.g., bearings, gearbox) unsupervised classification can be performed directly. On the other hand, for analysis where the features may be related to more than one fault (e.g., misalignment and mechanical looseness), the most relevant features (feature ranking) to identifying the sample as an anomaly are presented, allowing the specialist to analyze the problem in more detail, called root cause analysis.

The methodology is based on the feature importance ranking for each new sample identified as an anomaly, as presented in Algorithm \ref{alg:AlgorithmicLucas}. After identifying the anomaly in the previous part, the most relevant features are analyzed through the model's explanability. SHAP is used to obtain the feature importance ranking. The general features that only indicate the presence of a fault, but do not indicate the type / location are disregarded (e.g. rms and kurtosis). A new ranking of importance is obtained using only the specific features. For example, assuming that based on the importance score calculated by SHAP, the most relevant features in order are: rms, Ball Pass Frequency Outer (BPFO), Ball Pass Frequency Inner (BPFI), kurtosis, Ball Spin Frequency (BSF). Applying the methodology, the new ranking of importance would be: BPFO, BPFI and BSF. After that, according to the type of procedure applied, the result is obtained. For unsupervised classification, the specific features are analyzed, and the fault is classified based on the feature most relevant to the result. As each specific feature is related to a potential/unique type of component fault, the most relevant feature is considered as the fault present in the system. For root cause analysis, since the features may be related to more than one fault, the feature importance ranking is presented, assisting the specialist in identifying the type of fault.

Understanding which features the model uses to identify the anomaly is essential to perform root cause analysis/classification. In other words, through explainability it is possible to mimic human knowledge. Without the use of an explainability algorithm such as SHAP and Local-DIFFI, it is not possible to carry out the analysis, since the models used do not present explanations of how the final results were obtained. Thus, the association of state-of-the-art models to identify anomalies in the signals with algorithms that perform the explanability, allows the proposition of the new methodology.

Even though it is the state-of-the-art in explainability and model-agnostic, SHAP presents a high computational cost in relation to model-specific solutions. Therefore, a comparison was made using the recent proposed explainability algorithmic, Local-DIFFI for the Isolation Forest model. As stated above, the choice of model-specific Local-DIFFI is due to the fact that Isolation Forest presents excellent results in the literature and good robustness in relation to the variation of hyperparameters. Moreover, in general, Local-DIFFI presents very similar results to SHAP, as shown in \cite{carletti2020interpretable}. The similarity of the models was verified through Kendall-Tau rank distance, a metric commonly used for evaluation between two ranking lists.
\renewcommand{\baselinestretch}{0.5} 
\begin{algorithm}
\caption{Pseudo-Code}\label{euclid}
\label{alg:AlgorithmicLucas}
\begin{algorithmic}[1]
\Procedure{Unsupervised Classification}{}
\State $\textit{Type: specific analysis / specific feature related to a single fault}$ 
\State $\textit{Input: new sample}$ 
\State $\textit{Output: fault classification (most important specific feature)}$ 
\State
\If {$\textit{new sample = anomaly}$}
    \State $\textit{feature importance ranking} \gets \textit{shap or local-diffi(new sample)}$
    \EndIf
\State $\textit{feature importance ranking.}\textbf{drop}\textit{(general features)}$   
\State $\textit{feature importance ranking} \gets \textbf{sort}\textit{(feature importance ranking)}$
\State $\textit{most important feature} \gets \textit{feature importance ranking[0]}$
\State $\textit{print('The fault is located in:  ', most important feature)}$
\EndProcedure
\State
\Procedure{Root Cause Analysis}{}
\State $\textit{Type: general analysis / specific feature related to different faults}$ 
\State $\textit{Input: new sample}$ 
\State $\textit{Output: root causes (most important specific features)}$ 
\State
\If {$\textit{new sample = anomaly}$}
    \State $\textit{feature importance ranking} \gets \textit{shap or local-diffi(new sample)}$
    \EndIf
\State $\textit{feature importance ranking.}\textbf{drop}\textit{(general features)}$   
\State $\textit{feature importance ranking.} \gets \textbf{sort}\textit{(feature importance ranking)}$
\State $\textit{print('The root causes are related to:  ', feature importance ranking)}$
\EndProcedure
\end{algorithmic}
\end{algorithm}
\renewcommand{\baselinestretch}{1.5} 
\section{Experimental procedure}
\subsection{Data description}
Three datasets were used to address different faults found in rotating machinery. The faults analyzed were: defects in bearing and gearbox, misalignment, unbalance, mechanical looseness and combined faults. The use of different datasets, with different monitoring approaches, aims to validate the proposed methodology in different scenarios.

\subsubsection{Case 1: Bearing Dataset}

The first dataset considered (publicly available \cite{QIU20061066}), namely \emph{Bearing Dataset}, is composed by three run-to-failure tests with four bearings in each test. The rotation speed was kept constant at 2,000 rpm by an AC motor coupled to the shaft via rub belts. A radial load of 6,000 lb was applied to the shaft and bearing by a spring mechanism. Rexnord ZA-2115 double row bearings were installed on the shaft. PCB 353B33 accelerometers were installed on the bearings housing. All failures occurred after exceeding the projected bearing life, which is more than 100 million revolutions \cite{QIU20061066}. For the study, bearing 01 of test 02 was used. Each test consists of individual files of vibration signals recorded at specific intervals. Each file consists of 20,480 points with the sampling rate set at 20 kHz. NI DAQ Card 6062E was used for collection.

The dataset consists of run-to-failure tests, therefore no labels are available indicating the fault start: the only information provided is the type of fault present at the end of each test. To assess the efficiency of the AD model, the data was manually labeled. In the analysis, it was considered that after starting the defect, all subsequent observations correspond to a faulty bearing. It is worth mentioning that the labels were used only to evaluate the efficiency of the methodology and they were not used by the AD model.

The test has 984 observations, with the first 531 observations labeled as normal and the last 453 as anomalies (fault). The fault was identified in the outer race. The features used were: kurtosis, rms, BPFI, BPFO and BSF, which are widely used in bearing fault detection \cite{Canedo,ZHANG20112941,ZHANG20182426,LEI20091535,LI2017295,SINGH2019524}. Specific features are those that indicate the type of fault (BPFI, BPFO and BSF) and general features are those that indicate the presence of a defect (kurtosis and rms). The bearing fault frequencies are important to assess the type of defect and confirm its existence, which is not always noticed by other features. It is also important mentioning that there are cases where the fault does not present the classic defect behavior with the deterministic bearing frequencies in evidence \cite{SMITH2015100}, which makes it important to use other features. Knowing that bearing faults are generally associated with impacts, kurtosis is a relevant feature for the study. Finally, the rms value represents the global behavior of the system, indicating a general degradation and accentuation of the defect. The purpose of using this dataset, in addition to identifying the presence of the fault in a real monitoring situation, is to classify the type of fault using the proposed methodology.

\subsubsection{Case 2: Gearbox Dataset}

The second dataset considered, the \emph{Gearbox Dataset}, was presented in \cite{8360102} and it is used to evaluate faults in gearbox. A 32-tooth pinion and an 80-tooth gear were installed on the first stage input shaft. The second stage consists of a 48-tooth pinion and 64-tooth gear. The data were recorded using an accelerometer through a dSPACE DS1006 system, with sampling frequency of 20 KHz. Nine different gear conditions were introduced to the pinion on the input shaft, including healthy condition, missing tooth, root crack, spalling, and chipping tip with five different levels of severity. For each gear condition, 104 observations were collected resulting in a total of 936 observations. 

It is common knowledge that general gear problems tend to increase the energy of the sidebands spaced from the rotation frequency around the Gear Mesh Frequency (GMF) and their respective harmonics. Thus, simulating a real condition, the features used were: kurtosis, rms, 1xGMF, 2xGMF, 3xGMF, 4xGMF (1\textsuperscript{st}Stage), 1xGMF, 2xGMF (2\textsuperscript{nd} Stage). Due to non-stationary issues and the uncertainty caused by speed varying, instead of using the energy value in each GMF and respective side bands, the energy in the GMF band +/- 4*(nominal rotation frequency) was calculated. In addition to being able to detect the fault (AD), the use of this fault dataset aims to identify the location of the fault in the gearbox (first or second stage) and not to classify the type of fault (missing tooth, root crack, spalling and chipping tip).

\subsubsection{Case 3: Mechanical Fault Dataset}

The last dataset, \emph{Mechanical Fault Dataset}, was developed by one of the authors \cite{bd_pai_1,bd_pai_2}: the dataset contains different electrical and mechanical faults which were inserted in a experimental test rig; in this work we will consider the following faults: unbalance, misalignment, looseness and combined faults (being the combination of the previous ones). Six accelerometers were used to acquire the vibration signals, in the horizontal, vertical and axial positions, three in the fan-end side and three in the drive-end side.

The rotation speed was kept constant at 1717.5 rpm. The observations were labeled according to the fault introduced in the test rig, and later analysis of the vibration spectrum. Each file consists of 3,200 points with \textit{df} = 0.125 Hz. The dataset contains 5 conditions with a total of 1418 observations (532 normal, 557 unbalance, 283 misalignment, 28 mechanical looseness and 18 combined fault). 

In general, the unbalance is commonly identified in the vibration signal by increasing the energy in 1 x fr (speed rotation). It is noteworthy that other faults can also appear in 1 x fr as structural problems and even mechanical looseness. The most common types of misalignment and mechanical looseness show an increase in energy level in 2 x and 3 x fr, and therefore may have similar characteristics. The mechanical looseness can still have multiple and sub-harmonics of fr. Considering the types of faults and the respective behaviors, the following features were used: rms, energy level in 1 x fr, 2 x fr, 3 x fr and 4 x fr.

In addition to the basic objective of identifying the fault, the use of this dataset aims to evaluate the classification methodology with a focus on root cause analysis, when the features are correlated with more than one type of fault. The dataset also provides the possibility to study isolated and combined faults that, although known, have been little used in studies involving fault detection and new techniques of artificial intelligence compared to bearings and gearboxes.

\subsection{Analysis approaches }
Two approaches were used to define 3 different scenarios [Case 1, 2, 3] that can be found in real-world monitoring applications. 

In the first approach, a dynamic condition was considered with the data collected in sequence, where a temporal relationship and fault evolution is presented [Case 1]. For the study, a sliding window was used, where the training group was updated with each new sample, in case it was considered normal. 100 samples were initially used for the training group in order to ensure stability in the models. For this situation, as the model was started together with the machine under normal conditions (e.g.: after maintenance or a new machine), there are no anomalies in the training group. It is worth noting that this approach can also be used if there are anomalies in the training group (e.g.: cases of continuous monitoring where the machine was repaired after a fault, and it is desired to use all the signals to increase the amount of data in the model).

In the second approach, a static condition was considered, where the signals do not have a temporal correlation with each other [Case 2 and 3].  This approach simulates when historical data are available for the machine without labels. They also refer to different types of faults and normal conditions, however not necessarily collected in sequence. It is important to highlight that although Cases 2 and 3 represent the same condition called static condition, the types of faults studied are different in each case. The data were divided into training and test groups. Due to the number of observations available, the size of the training group is limited by the number of normal samples and the rest designated as a test. The training group consisted of 80\% of samples of normal condition and 20\% of anomalies selected at random. The proportion has been defined as a machine in operation is mostly in normal condition, and few situations with faults. Such an approach also shows that it is possible to implement the proposed methodology even with the presence of anomalies in training set.

\subsection{Hyperparameter tuning}

The hyperparameters for each model were adjusted based on the training group to obtain the best performance and are shown in Table~\ref{tab:hyperparameter}. The hyperparameters are presented in relation to the library used \cite{zhao2019pyod}. As the models did not show significant differences in the final result in relation to the hyperparameters for each case, the hyperparameters were kept the same for all analysis.
\renewcommand{\baselinestretch}{1.2} 
\begin{table*}[ht]
   \caption{}
   \caption*{Hyperparameter for each model}\label{tab:hyp}
   \label{tab:hyperparameter}
   \resizebox{\textwidth}{!}{\begin{tabular}{ll}
     \toprule
       Model and Hyperparameter \\
     \midrule
        kNN & n\_neighbors=5, method=largest, metric='minkowski' \\
        MCD & assume\_centered=False \\
        LOF & n\_neighbors=16 \\
        CBLOF & n\_clusters=6, alpha=0.8, beta=4 \\
        OCSVM & kernel='rbf', gamma=0.2, nu=0.7 \\
        FB & base\_estimator=LOF, n\_estimators=10, max\_features=1.0, combination='average' \\
        FastABOD & n\_neighbors=5 \\
        IF & n\_estimators=100, max\_samples=128 \\
        HBOS & n\_bins=5, alpha=0.1, tol=0.5 \\
        LODA & n\_bins=5, n\_random\_cuts=50 \\
     \bottomrule
   \end{tabular}}
\end{table*}
\renewcommand{\baselinestretch}{1.5} 
\subsection{Evaluation metrics}

For the fault detection part, as an unsupervised methodology, at the end of the test the anomaly score is calculated, where samples with high anomaly score values are usually anomalies. To verify the performance of the proposed methodology, threshold values were defined based on the training group. For the bearing dataset (Case 1) the threshold was defined based on the assumption that the training group is composed of only signals in normal condition (considering that the initial signals correspond to the start of operation of the bearing). As the gearbox dataset (Case 2) and mechanical fault (Case 3), the contamination ratio is known, its value was used to define the threshold. It is also worth mentioning that, due to the knowledge about the fault characteristics and respective behavior, the user can adjust the contamination rate of the methodology during the application, based on a preliminary analysis of the training data.

For the static condition, each test was performed 100 times to show the stability of the model. The signals were randomly chosen for the test and training group at each iteration of the model. For the dynamic condition, in each new update of the training group, 5\% of the samples were randomly excluded to also assess the stability of the model. As the update occurred more than 400 times in the tested dataset, the complete test was performed 10 times. In addition to the variation of the dataset, each iteration of the model was performed with different random seeds.

The results are presented using the F1-Score, PR-AUC (Precision-Recall Area Under the Curve) and average confusion matrix of the iterations with respective standard deviations. The metrics were chosen due to the greater interest in correctly identifying samples referring to faults (anomalies). Although it is a problem to have false positives in the final result, failing to acknowledge a fault is even worse as it can result in the machine breakdown. Moreover, these metrics are also important when dealing with unbalanced dataset (common situation in the real scenario). 

For the fault diagnosis using the unsupervised classification approach (Case 1 and 2), each sample identified as anomaly is classified in relation to the type / location of the fault. As the classification was performed only for the anomalies identified, accuracy was used as an evaluation metric. For the root cause analysis (Case 3), the feature importance ranking is presented. Kendall Tau distance was used to compare SHAP and Local-DIFFI. The tests were performed using 2.2 GHz Intel Core i7 Dual-Core, 8 GB 1600 MHz DDR3, Intel HD Graphics 6000 1536 MB.

\section{Results and discussion}
\subsection{Data Exploration}
In this subsection the data used in this work for Case 1-3 are analyzed and discussed.
\subsubsection{Case 1: Bearing Dataset}

\begin{figure}%
    \centering
    \subfloat[\centering Complete signal for the test in time domain\label{fig:time_bearing}]{{\includegraphics[width=9cm]{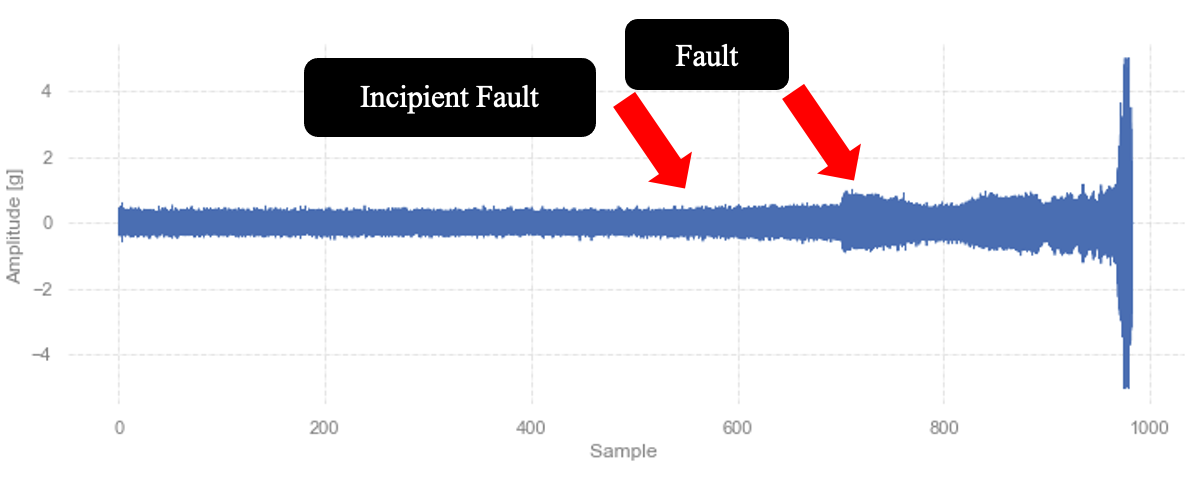} }}%
    \qquad
    \subfloat[\centering Waterfall envelope spectrum\label{fig:cascata}]{{\includegraphics[width=6.5cm]{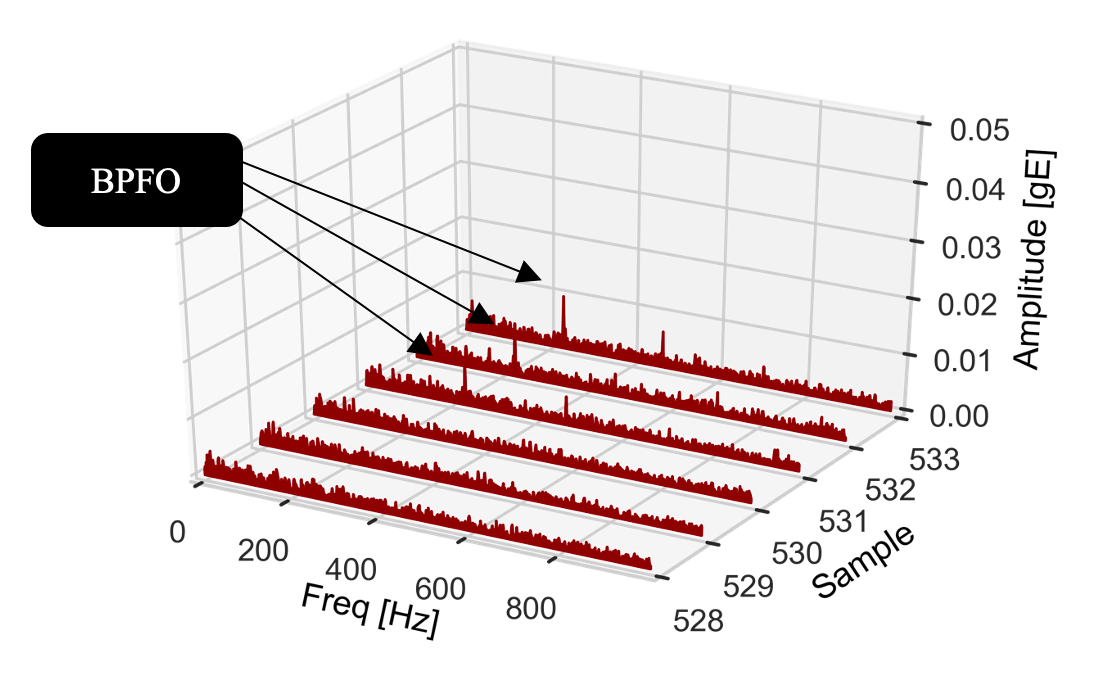} }}%
    \caption{Bearing dataset signal.}%
    \label{fig:sinalcompleto}%
\end{figure}

Fig. \ref{fig:time_bearing} shows the complete signal for the test in time domain. As the signal was not collected continuously (24/7), it was decided to present it according to the sample (x-axis). The point at the incipient fault starts, as well as the fault are identified.
\renewcommand{\baselinestretch}{1.3} 
\renewcommand{\baselinestretch}{1.5} 
In Fig. \ref{fig:time_bearing} it can be seen that although the fault is easily identified by the signal trend in the time domain, the incipient fault is not easily identified by visual analysis. Making it important to use the ML model with the appropriate features to provide the maintenance team adequate time to schedule an intervention. Fig. \ref{fig:cascata} shows the moment of beginning of the incipient fault presented in the envelope spectrum and used to define the labels of the signals. It is noted that from the sample 531 there is evidence of BPFO, being defined as indicative of incipient fault and, therefore, anomaly. Based on the adopted methodology, all samples after this signal are considered faults (anomalies).

\begin{figure}[ht]
  \subfloat[Normal]{
	\begin{minipage}[c][0.65\width]{
	   0.3\textwidth}
	   \centering
	   \includegraphics[width=1\textwidth]{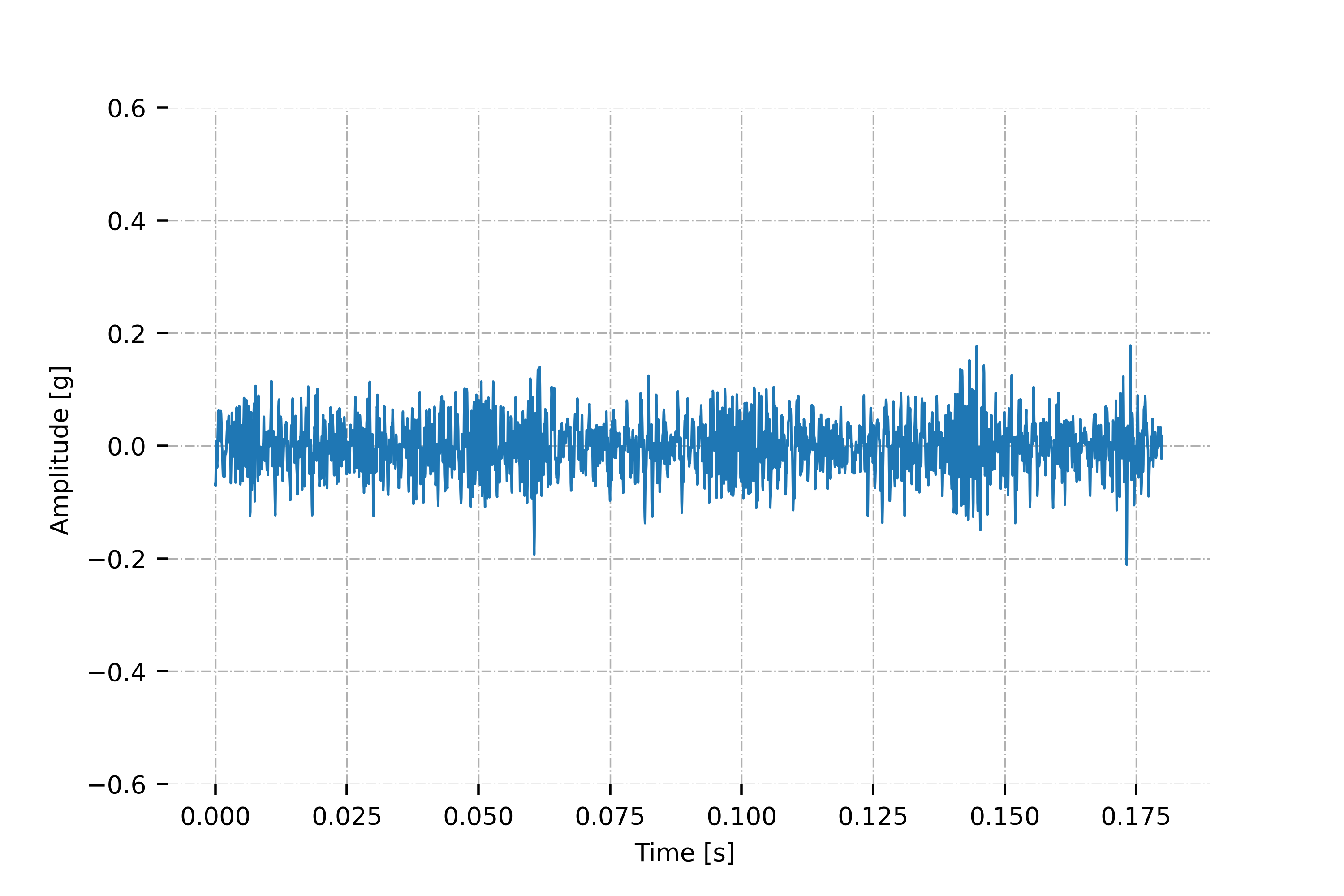}
	\end{minipage}}
 \hfill 	
  \subfloat[Root crack]{
	\begin{minipage}[c][0.65\width]{
	   0.3\textwidth}
	   \centering
	   \includegraphics[width=1\textwidth]{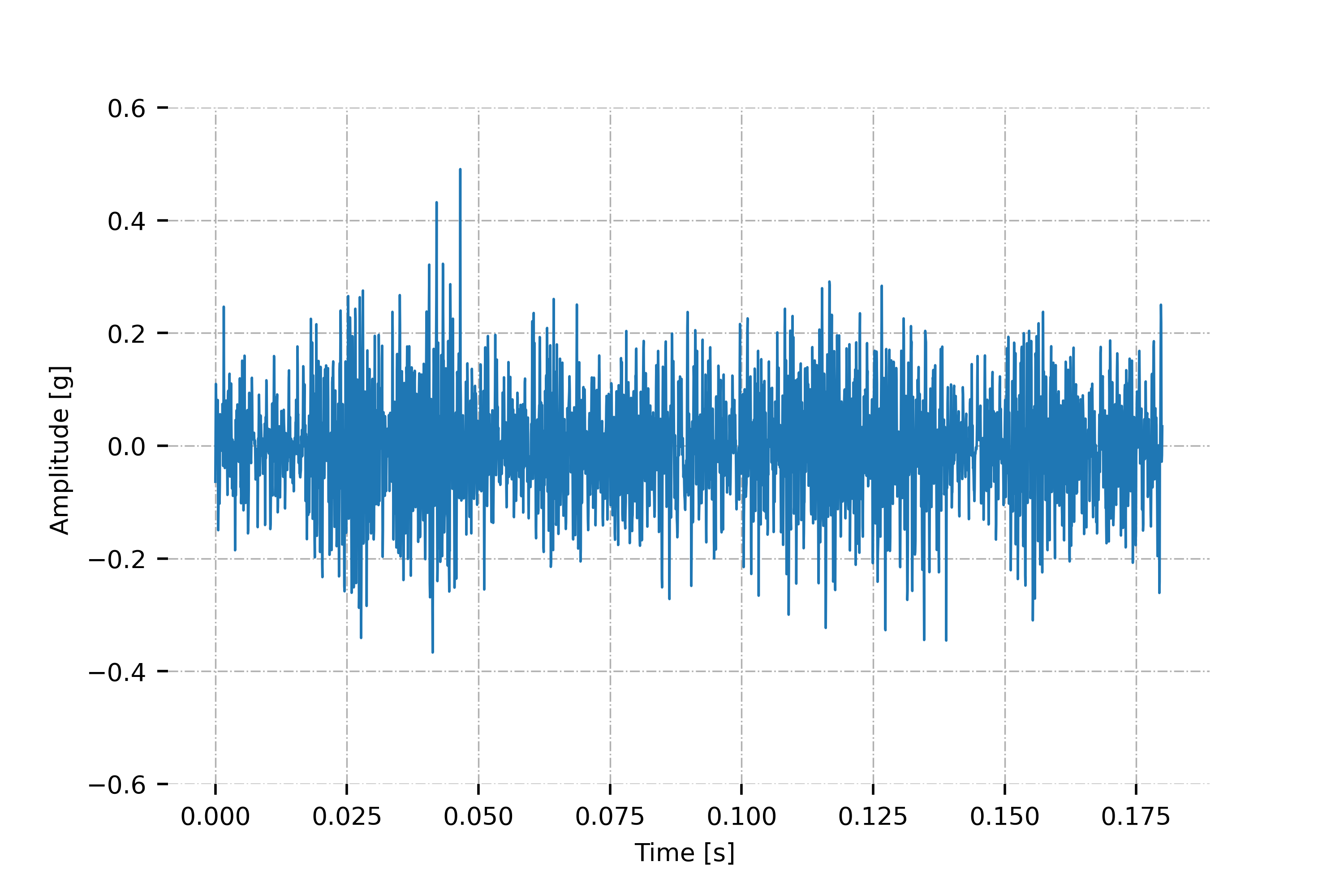}
	\end{minipage}}
 \hfill	
  \subfloat[Missing Tooth]{
	\begin{minipage}[c][0.65\width]{
	   0.3\textwidth}
	   \centering
	   \includegraphics[width=1\textwidth]{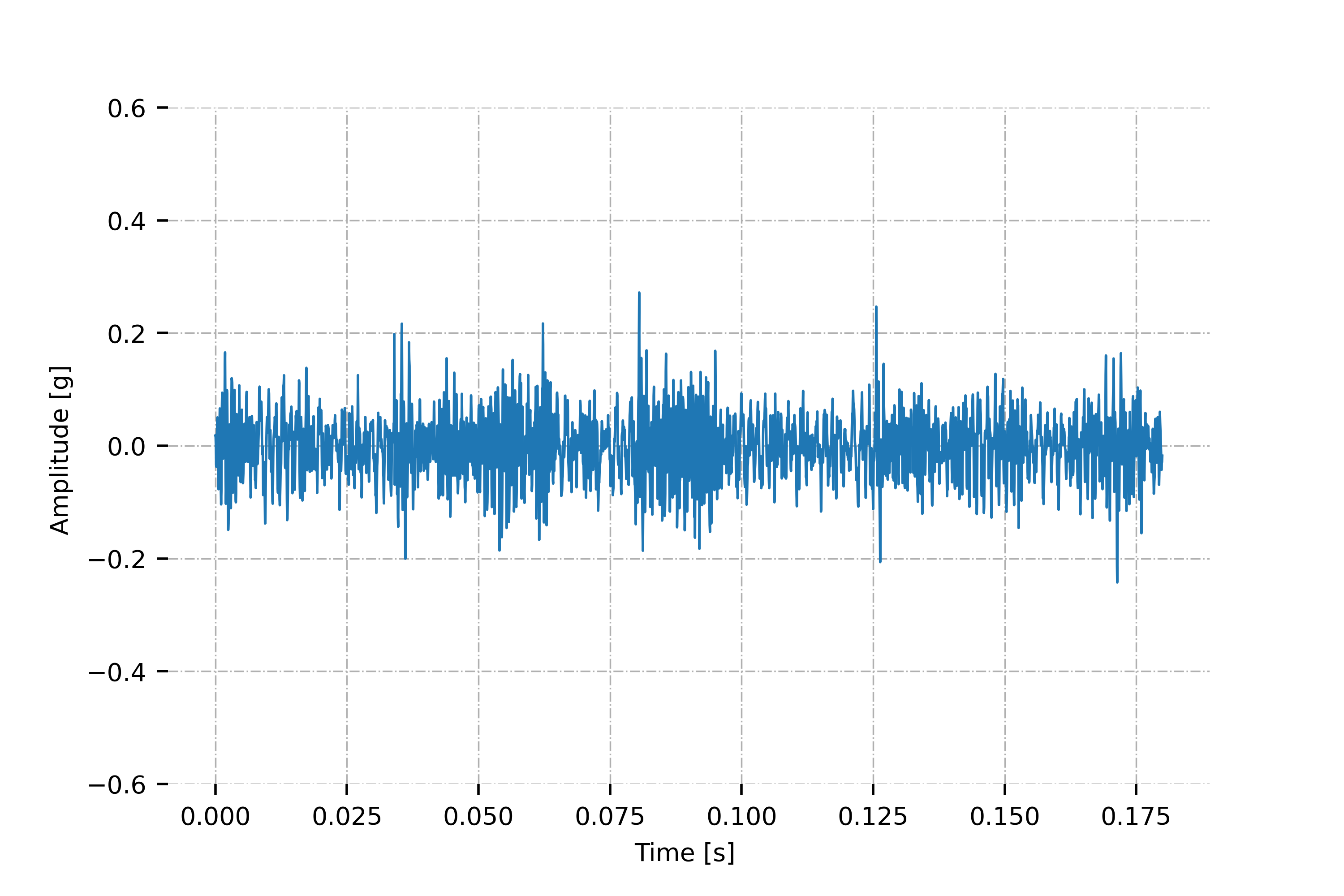}
	\end{minipage}}\\
  \subfloat[Spalling]{
	\begin{minipage}[c][0.65\width]{
	   0.3\textwidth}
	   \centering
	   \includegraphics[width=1\textwidth]{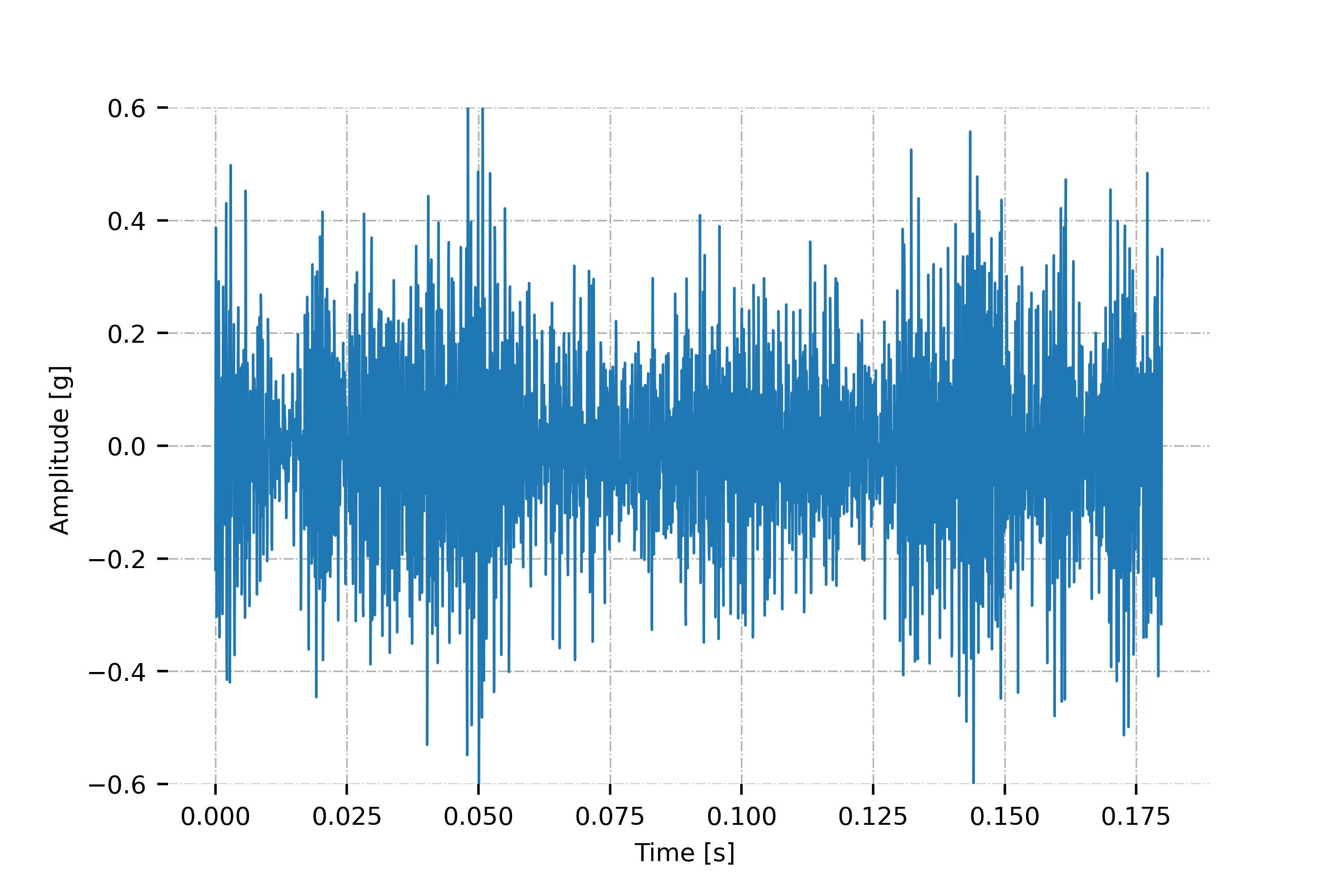}
	\end{minipage}}
 \hfill 	
  \subfloat[Chipping tip (least severe)]{
	\begin{minipage}[c][0.65\width]{
	   0.3\textwidth}
	   \centering
	   \includegraphics[width=1\textwidth]{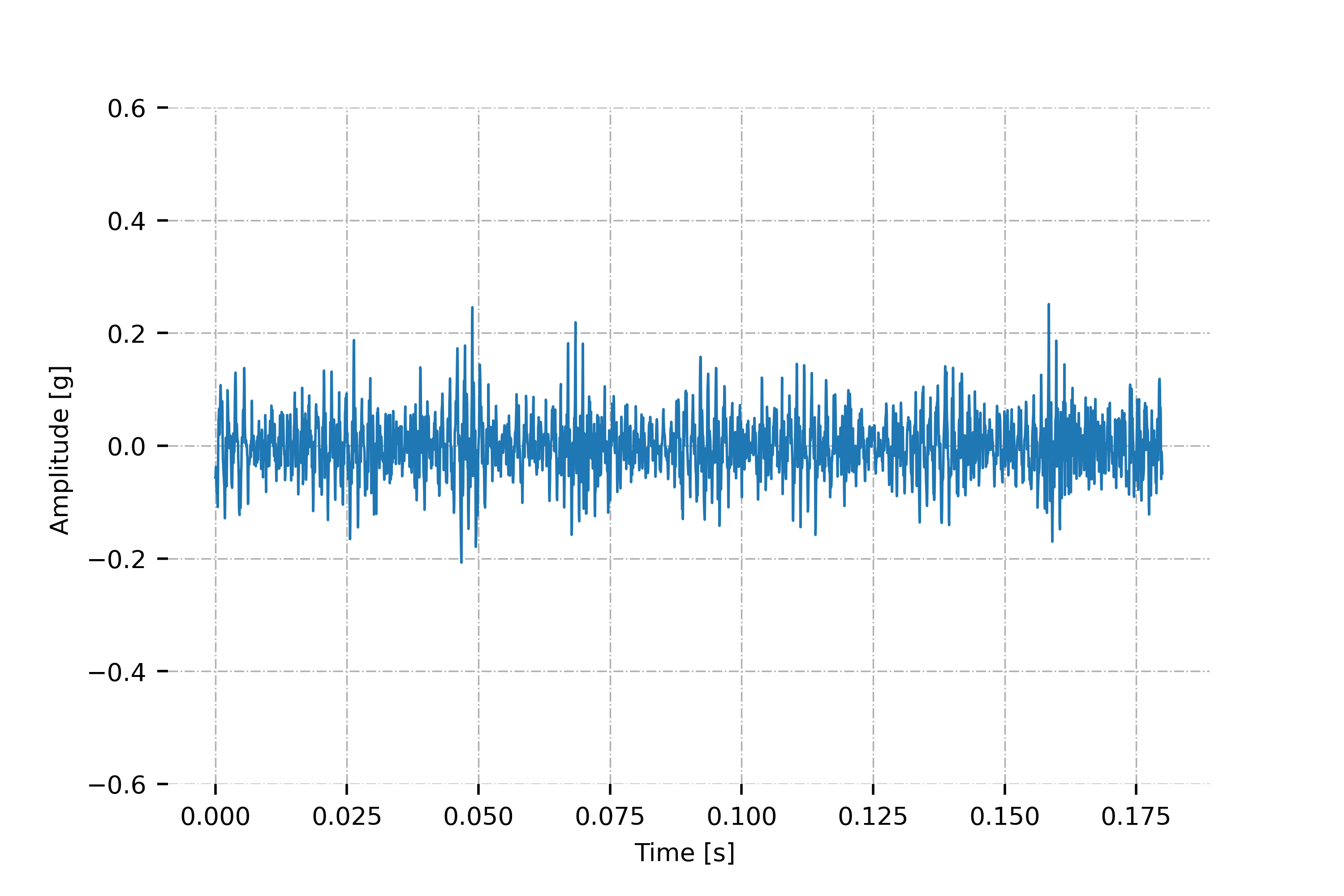}
	\end{minipage}}
 \hfill	
  \subfloat[Chipping tip (most severe)]{
	\begin{minipage}[c][0.65\width]{
	   0.3\textwidth}
	   \centering
	   \includegraphics[width=1\textwidth]{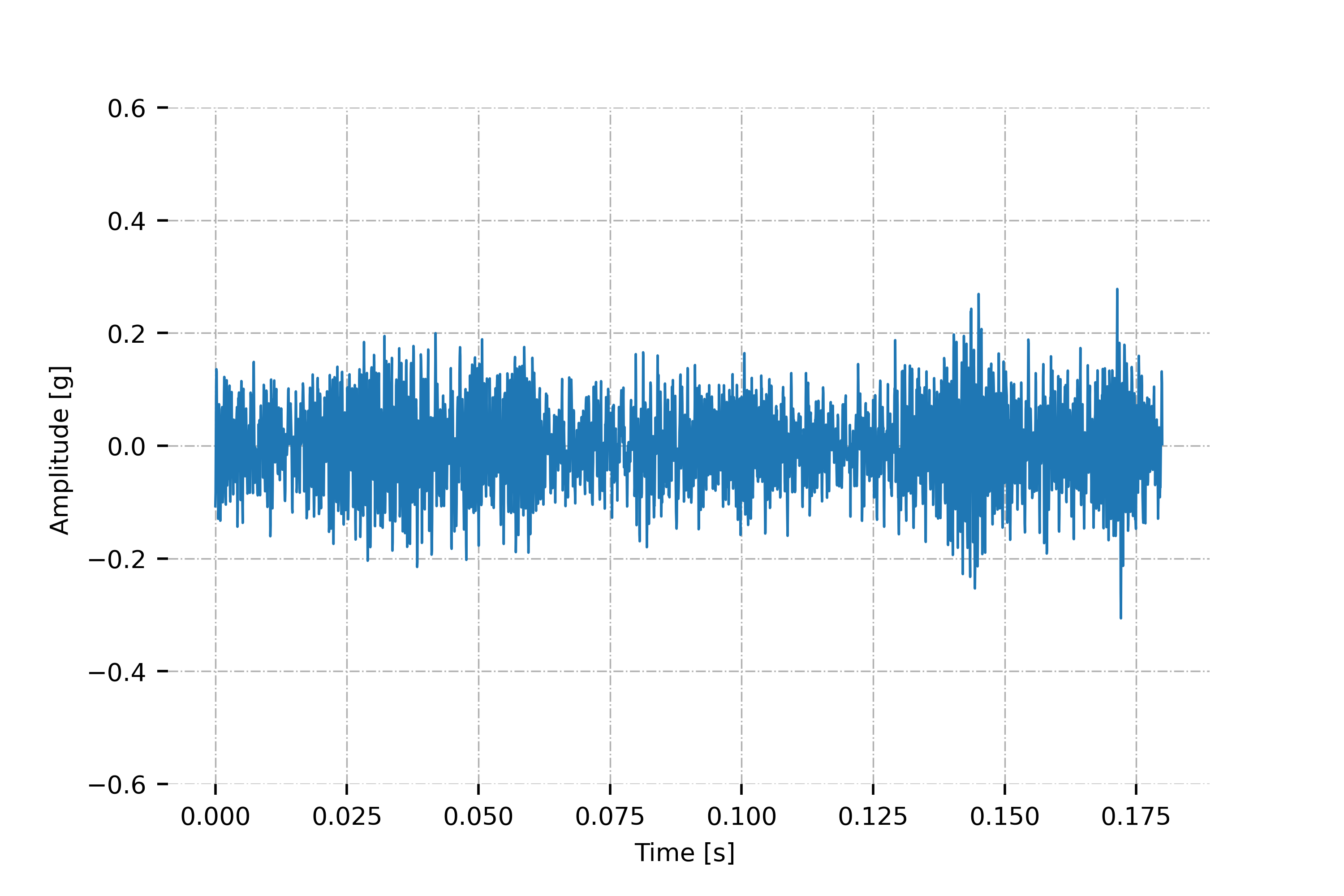}
	\end{minipage}}
    \caption{Vibration signal examples under different gear health conditions.}
    \label{fig:sinalgearbox}
\end{figure}

The signals for the different types of faults present in the dataset are shown in Fig. \ref{fig:sinalgearbox}. Due to the possibility of non-stationarity caused by the variation of the load, it was decided to show the signals in the time domain. In addition, some defects, such as broken / cracked tooth, can also be better viewed.

It is possible to notice an increase in the energy level in the signal for defects such as root crack, spalling and chipping tip (most severe). On the other hand, differentiating a normal signal from one with a missing tooth or chipped tip in the initial stage is not so simple. Therefore, the feature extraction and the use of artificial intelligence techniques become essential for more assertive monitoring.
\renewcommand{\baselinestretch}{1.3} 
\renewcommand{\baselinestretch}{1.5} 
\subsubsection{Case 3: Mechanical Faults}

\begin{figure}[ht]
  \subfloat[Normal]{
	\begin{minipage}[c][0.65\width]{
	   0.3\textwidth}
	   \centering
	   \includegraphics[width=1\textwidth]{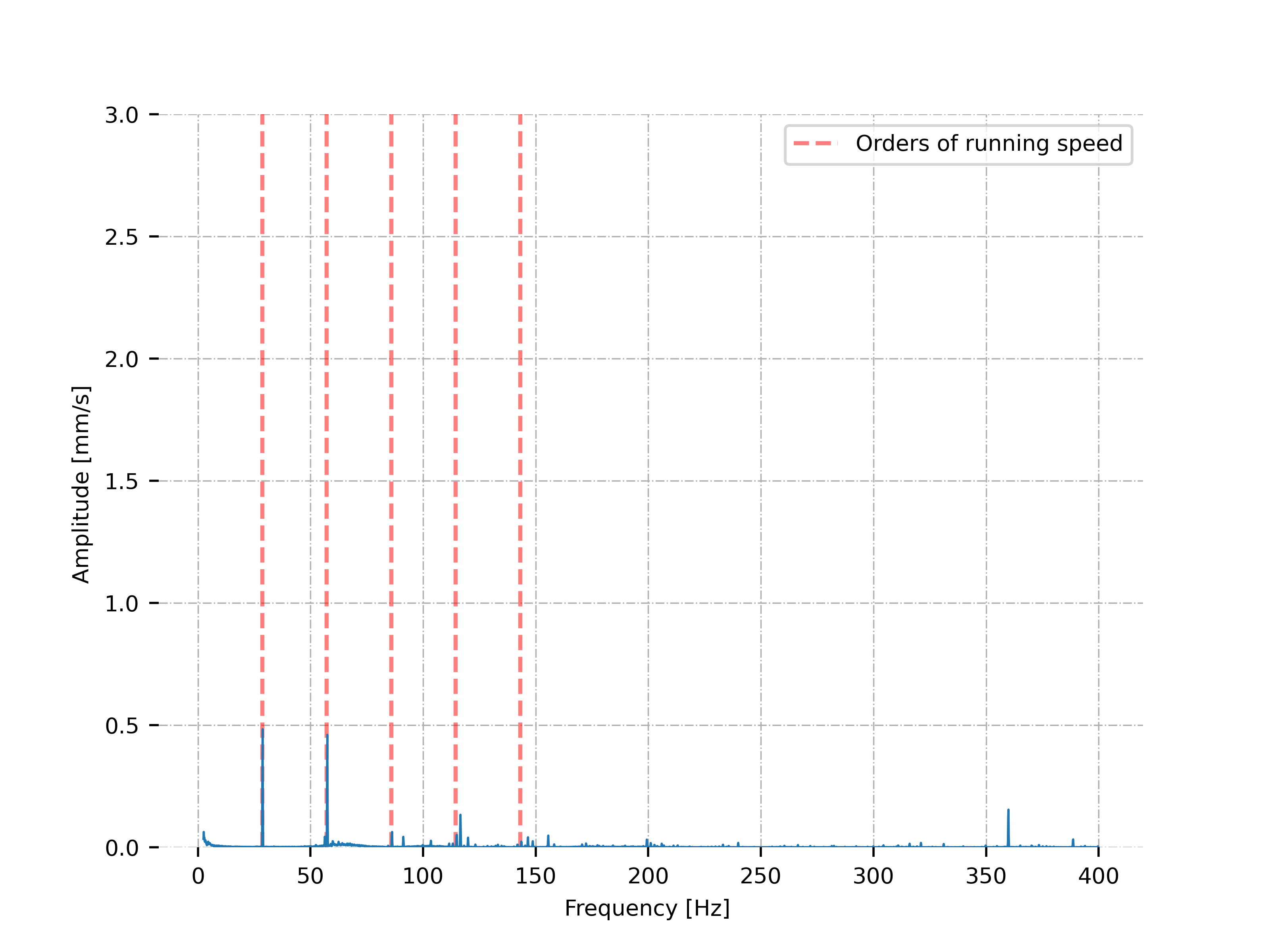}
	\end{minipage}}
 \hfill 	
  \subfloat[Unbalance]{
	\begin{minipage}[c][0.65\width]{
	   0.3\textwidth}
	   \centering
	   \includegraphics[width=1\textwidth]{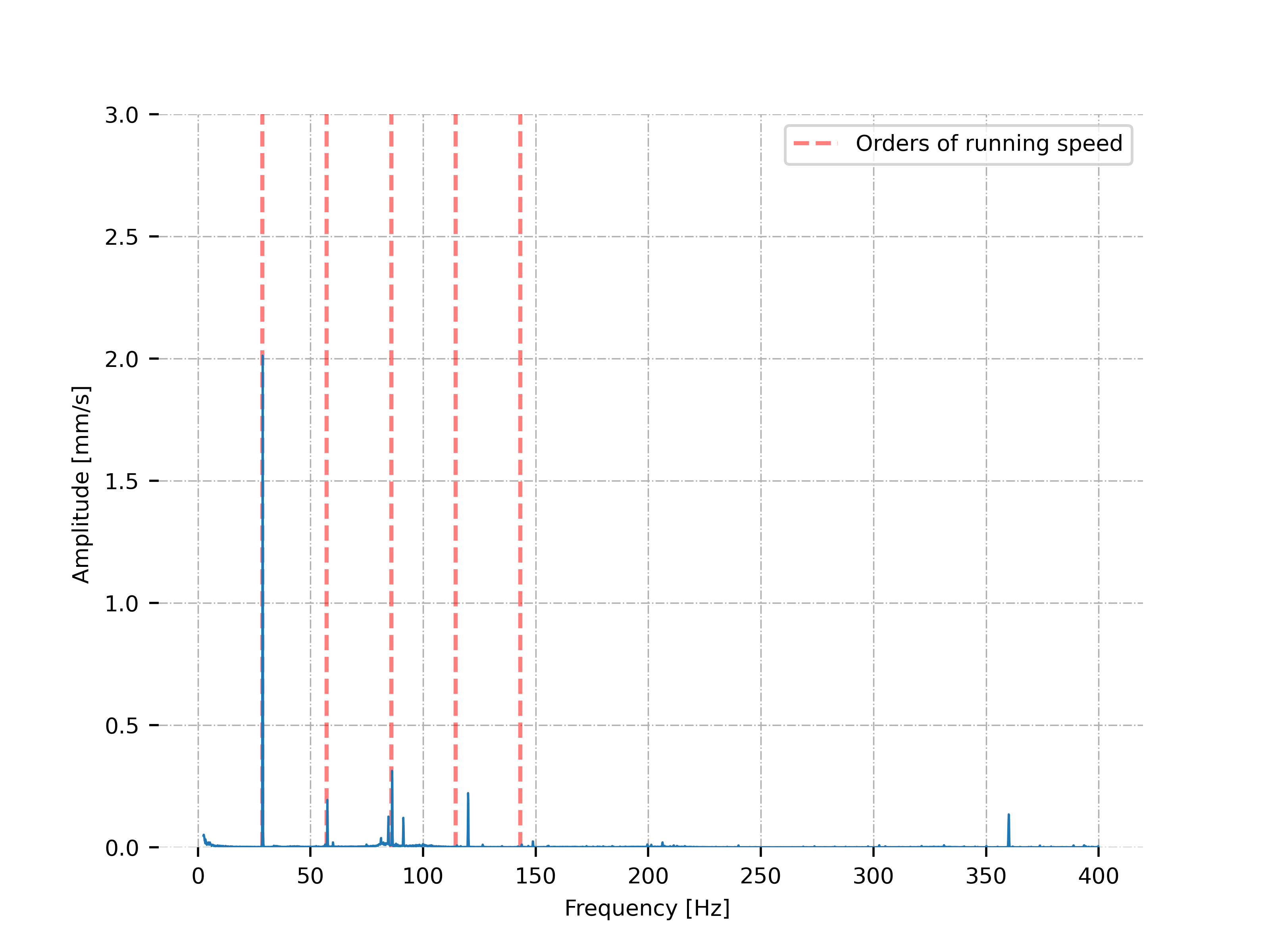}
	\end{minipage}}
 \hfill	
  \subfloat[Misalignment]{
	\begin{minipage}[c][0.65\width]{
	   0.3\textwidth}
	   \centering
	   \includegraphics[width=1\textwidth]{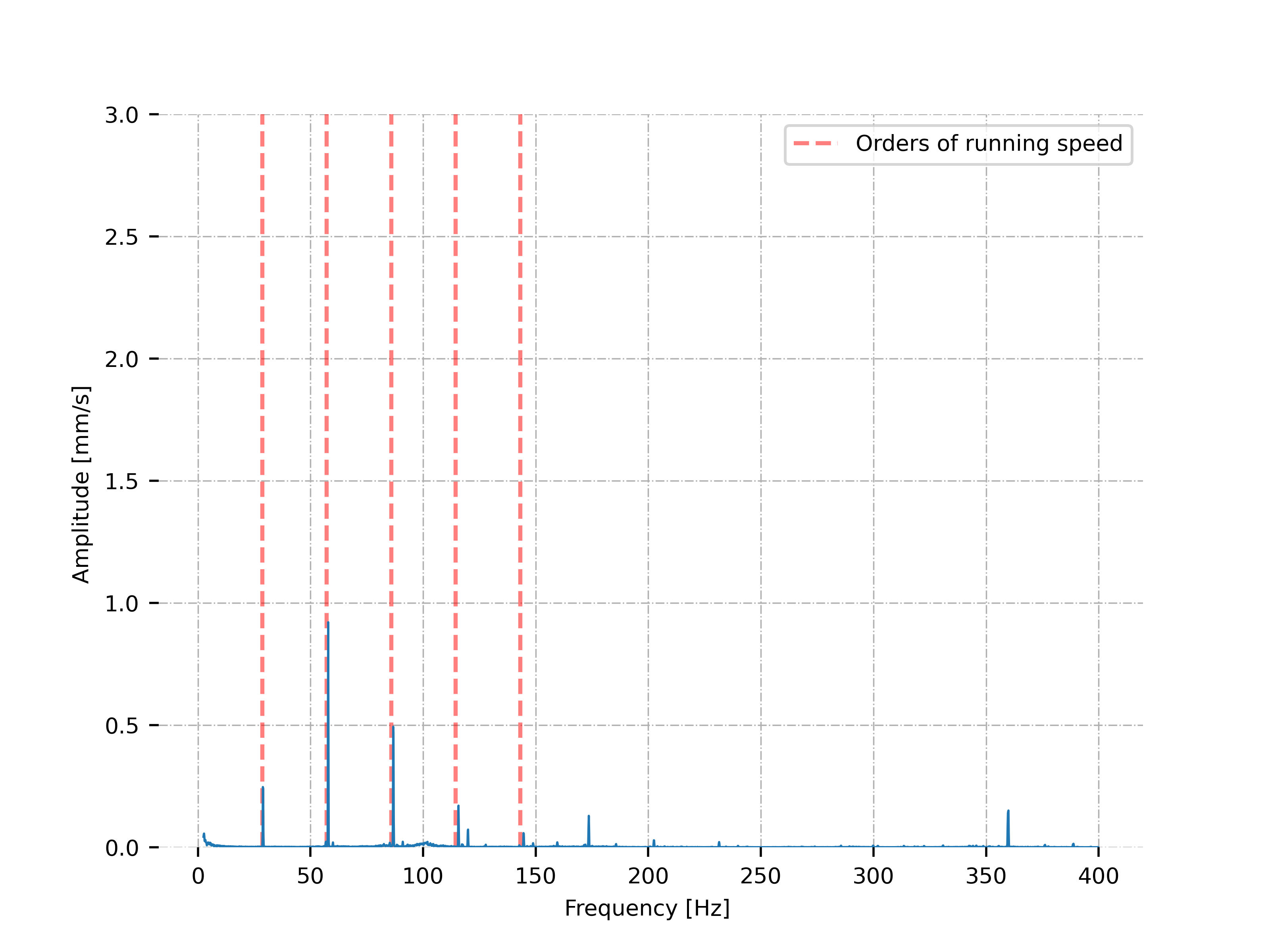}
	\end{minipage}}\\
  \subfloat[Looseness]{
	\begin{minipage}[c][0.45\width]{
	   0.5\textwidth}
	   \centering
	   \includegraphics[width=0.6\textwidth]{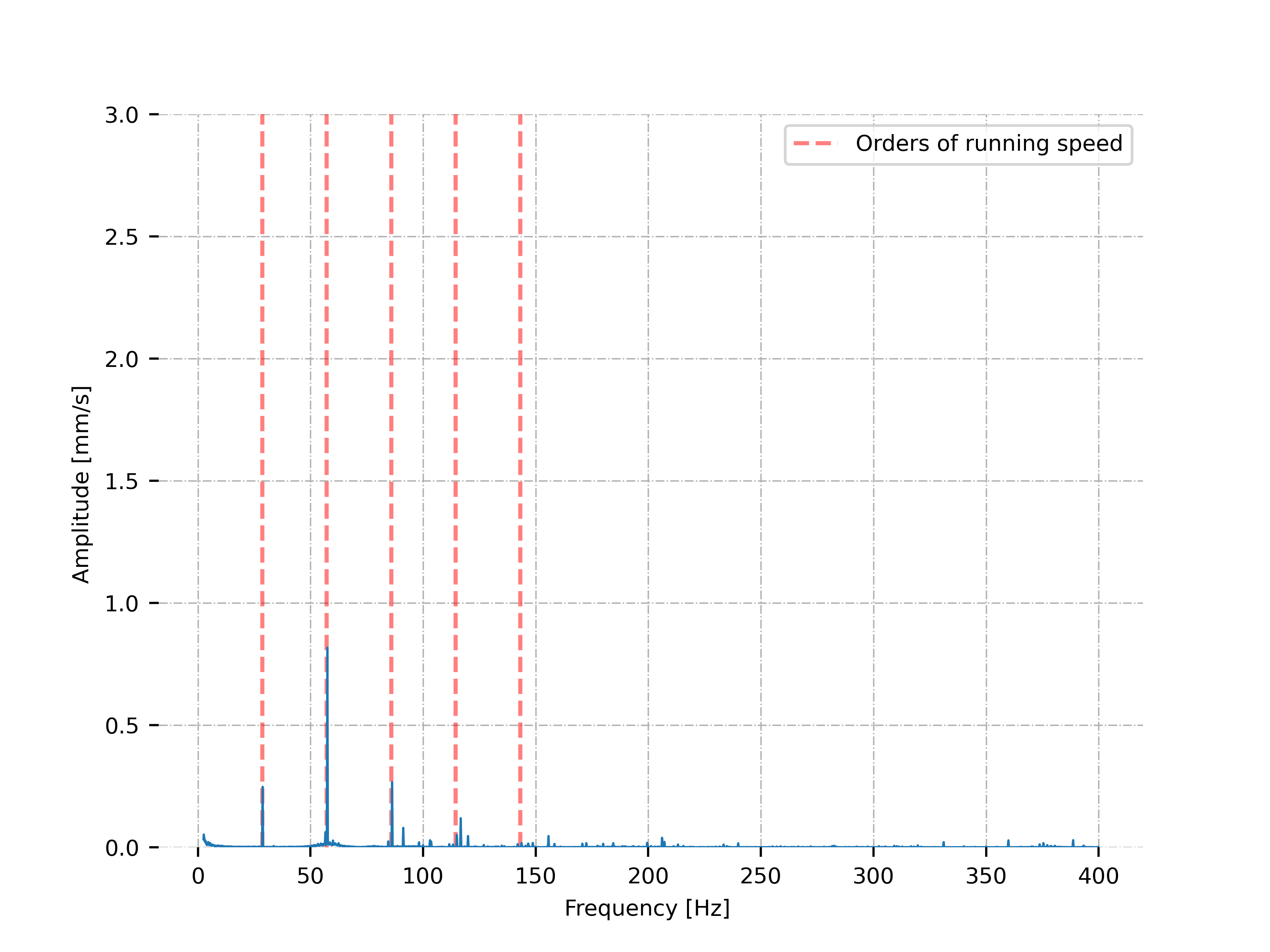}
	\end{minipage}}
 \hfill 	
  \subfloat[Combined Faults]{
	\begin{minipage}[c][0.45\width]{
	   0.5\textwidth}
	   \centering
	   \includegraphics[width=0.6\textwidth]{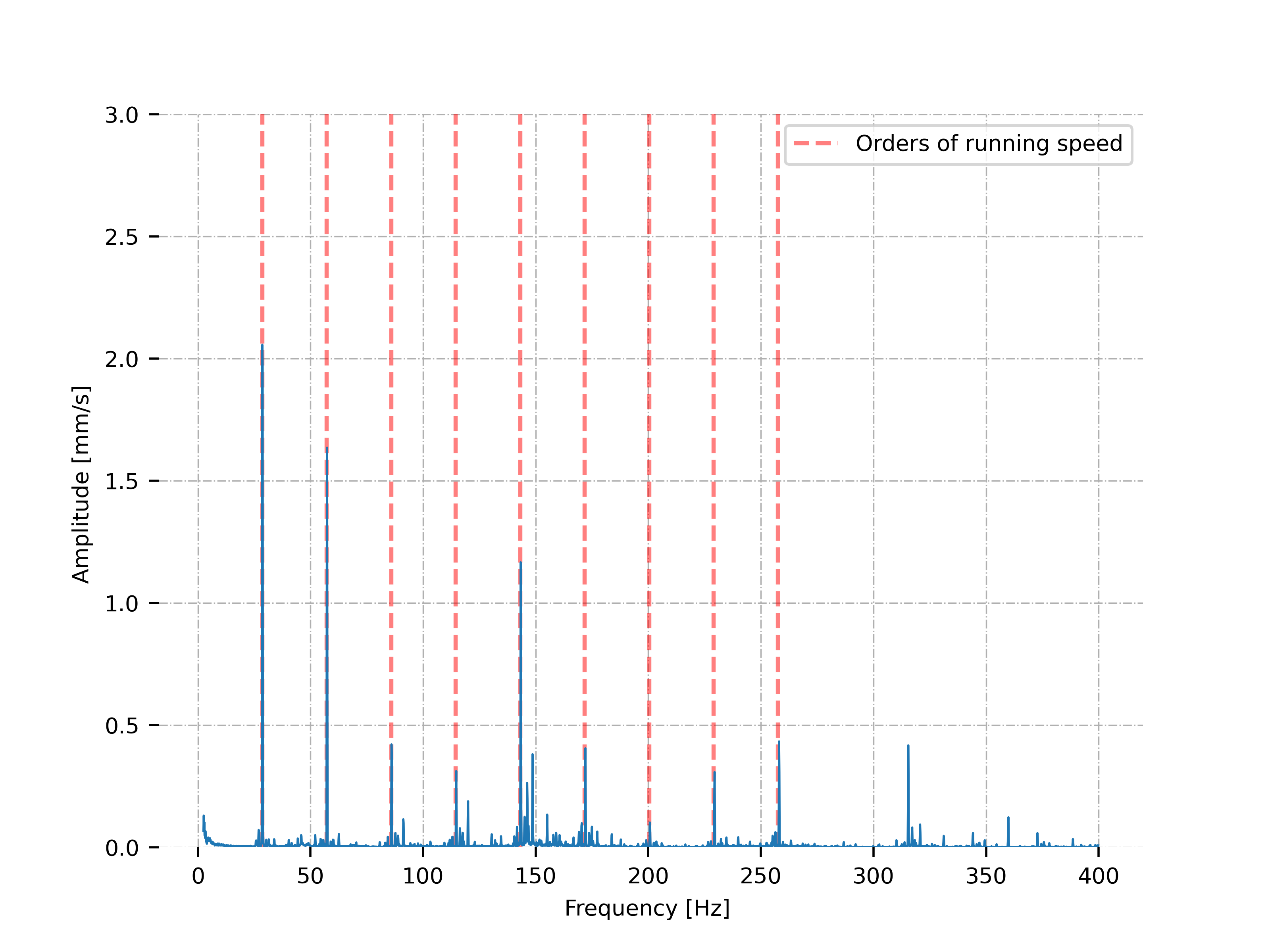}
	\end{minipage}}
    \caption{Examples of vibration signals for different faults present in the dataset.}
    \label{fig:sinalmechfaults}	
\end{figure}

In Fig. \ref{fig:sinalmechfaults} some examples of faults are shown. The frequency domain was used to better characterize the faults, knowing that the speed rotation was kept constant.

For the normal situation, there is no predominance of any characteristic frequency, in addition to presenting a low level of vibration in relation to other situations. In the unbalance case, it is evident the increase in energy in 1 x fr, characteristic of the fault. Misalignment and mechanical looseness exhibit very similar behavior in the signal with 2 x fr greater than the other harmonics. The differentiation was performed based on the type of fault inserted in the test rig. For the situation of combined failures (unbalance, misalignment and mechanical looseness) the characteristics of all faults are noted.

For the reasons mentioned above, the classification of such faults includes the analysis of signals in other positions and complementary techniques. Therefore, for this case, the proposed classification methodology will provide only the most relevant features for the identification of the fault, assisting the specialists in the search for the root cause of the problem.
\renewcommand{\baselinestretch}{1.3} 
\renewcommand{\baselinestretch}{1.5} 
\subsection{Fault detection: Anomaly Detection }

Using the proposed methodology, the results obtained for the fault detection are presented in Table~\ref{tab:fault_detection}. Table~\ref{tab:fault_detection} also shows the average time spent for training and testing (a new sample). The top three results for each metric are shown in bold. It can be seen in Table~\ref{tab:fault_detection} for Case 1 and 3, that the models that had better identification of the faults through the F1-Score were: MCD, HBOS and IF. For Case 2, although HBOS had a very close performance, the models that showed the best results were: MCD, kNN and IF.
\renewcommand{\baselinestretch}{1.3} 
\begin{table*}[ht]
   \caption{}
   \caption*{Fault detection results}
   \label{tab:fault_detection}
   \resizebox{\textwidth}{!}{\begin{tabular}{cccccccccccc}
     \toprule
     Metric & kNN & MCD & LOF & CBLOF & OCSVM & FB & FastABOD & IF & HBOS & LODA & Ensemble\\
     \midrule
     \textbf{Case 1} \\
     F1-Score & 63.21 & \textbf{99.45} & 57.94 & 60.87 & 60.78 & 61.99 & 88.17 & \textbf{97.19} & \textbf{98.43} & 39.12 & 70.20\\
      & (1.04) & (0.03) & (0.71) & (0.48) & (0.17) & (2.93) & (1.34) & (1.06) & (0.79) & (10.82) & (1.40)\\
     PR AUC & 96.45 & \textbf{99.91} & 92.02 & 93.55 & 93.28 & 94.35 & 98.80 & \textbf{99.92} & \textbf{99.91} & 94.23 & 98.85\\
     & (0.04) & (0.00) & (0.12) & (0.21) & (0.02) & (0.19) & (0.08) & (0.01) & (0.01) & (0.81) & (0.08)\\ 
     Time [s] & \textbf{0.0041} & 0.3066 & \textbf{0.0058} & 0.0835 & 0.0101 & 0.0555 & 0.1081 & 0.3671 & \textbf{0.0049} & 0.0307 & 1.3435\\
     
     \textbf{Case 2} \\
     F1-Score & 99.82 & \textbf{99.84} & 99.26 & 99.06 & 89.68 & 99.64 & 97.16 & \textbf{99.71} & 99.49 & 90.14 & 99.70\\
      & (0.08) & (0.08) & (0.39) & (0.69) & (2.21) & (0.20) & (1.80) & (0.21) & (0.18) & (6.42) & (0.05)\\
     PR-AUC & 99.97 & \textbf{99.99} & 99.98 & 99.95 & 99.74 & \textbf{99.99} & 99.88 & \textbf{99.99} & 99.95 & 99.83 & \textbf{99.99}\\
     & (0.02) & (0.00) & (0.01) & (0.04) & (0.71) & (0.01) & (0.12) & (0.01) & (0.04) & (0.11) & 0.01\\
     Time [s] & 0.1706 & 0.0807 & \textbf{0.0165} & 0.0575 & \textbf{0.0073} & 0.1119 & 1.0303 & 0.4121 & \textbf{0.0101} & 0.0434 & 1.9404\\
     
     \textbf{Case 3} \\
     F1-Score & 96.27 & \textbf{98.15} & 92.01 & 94.83 & 95.62 & 92.35 & 95.76 & \textbf{97.20} & \textbf{99.22} & 97.16 & 97.10\\
      & (0.00) & (0.01) & (0.00) & (0.15) & (0.00) & (0.33) & (0.00) & (0.27) & (0.00) & (0.39) & (0.12)\\
     PR-AUC & 99.60 &\textbf{99.91} & 98.76 & 99.38 & 99.50 & 98.83 & 99.51 & \textbf{99.74} & \textbf{99.96} & 99.69 & 99.53\\
     & (0.00) & (0.00) & (0.00) & (0.03) & (0.00) & (0.04) & (0.00) & (0.05) & (0.00) & (0.20) & (0.00)\\
     Time [s] & 0.1721 & 0.6172 & \textbf{0.0251} & 0.1030 & \textbf{0.0209} & 0.1808 & 0.6227 & 0.4195 & \textbf{0.0361} & 0.0411 & 2.2384\\
     \bottomrule
   \end{tabular}}
\end{table*}
\renewcommand{\baselinestretch}{1.5} 

In order to evaluate the general efficiency of the model, regardless of the defined threshold, the PR-AUC value was calculated. The results show that by modifying the threshold, the models can present even better results. It is noteworthy that the threshold used for comparison and calculation of the F1-Score was defined based on the previous analysis of the training group, simulating a real condition where the data for testing are not yet available. For this reason, it was decided to present the F1-Score value based on the defined threshold instead of the optimum value that could be obtained by adjusting the threshold in the complete dataset. Nevertheless, it can also be analyzed that, despite the improvement in the results, in general, the models that showed better performance in relation to PR-AUC were the same ones with highest F1-Score value: IF, HBOS and MCD. 

Although in general HBOS, MCD and IF presented good results for the three cases, it can be seen that depending on the dataset, other models can obtain better performance, such as kNN in Case 1 and 2. The good results obtained in Table~\ref{tab:fault_detection} for the three cases show that it is possible to detect faults in rotating machinery through the models studied in an unsupervised way. 

Among the models with the best results, HBOS presented the lowest computational time. In general, LOF and OCSVM also presented low values. On the other hand, FastABOD, MCD and IF demanded more computational time in relation to the other models (in a general analysis, excluding Ensemble). The low average time for most models in training and testing a sample allow implementation in an industrial environment focused on predictive maintenance. 

For the proposed comparison between SHAP and Local-DIFFI, and due to the good overall performance of Isolation Forest, the details of the methodology results are presented for the model. The average values for the confusion matrix are presented in Table~\ref{tab:confusionmatrix} (the sample quantities were rounded up because they are integer values). The confusion matrix allows a better visualization of the results in relation to the distribution of the signals in the respective classes. The results are presented both in percentage and in quantity of signals.
\renewcommand{\baselinestretch}{1.3} 
\begin{table*}[ht]
   \caption{}
   \caption*{Confusion Matrix}
   \label{tab:confusionmatrix}
   \resizebox{\textwidth}{!}{\begin{tabular}{ccccccccc}
     \toprule
    \textbf{Case 1} & Normal$^2$ & Fault$^2$ & \textbf{Case 2} & Normal$^2$ & Fault$^2$ & \textbf{Case 3} & Normal$^2$ & Fault$^2$ \\
     \midrule
     \multirow{2}{*}{Normal$^1$}  
        & \multicolumn{1}{c}{48.75 \% (0\%)} & \multicolumn{1}{c}{0 \% (0\%)} & \multirow{2}{*}{Normal$^1$} 
        & \multicolumn{1}{c}{2.63 \% (0.23\%) } & \multicolumn{1}{c}{0.24 \% (0.11\%) } &
         \multirow{2}{*}{Normal$^1$}
        & \multicolumn{1}{c}{10.05 \% (0.36\%)} & \multicolumn{1}{c}{3.06 \% (0.36\%)}  \\
        & \multicolumn{1}{c}{431 (0) } & \multicolumn{1}{c}{0 (0)} & & \multicolumn{1}{c}{22 (2)} & \multicolumn{1}{c}{2 (1)} & & \multicolumn{1}{c}{82 (3)} & \multicolumn{1}{c}{25 (3)} \\
     
     \multirow{2}{*}{Fault$^1$}  
        & \multicolumn{1}{c}{2.82 \% (1.13\%)} & \multicolumn{1}{c}{48.41 \% (1.01\%)} & \multirow{2}{*}{Fault$^1$} 
        & \multicolumn{1}{c}{0.24 \% (0.23\%) } & \multicolumn{1}{c}{96.89 \% (0.35\%)} &
         \multirow{2}{*}{Fault$^1$}
        & \multicolumn{1}{c}{1.84 \% (0.24\%)} & \multicolumn{1}{c}{85.05 \% (0.24\%)}  \\
        & \multicolumn{1}{c}{25 (10)} & \multicolumn{1}{c}{428 (9)} & & \multicolumn{1}{c}{2 (2)} & \multicolumn{1}{c}{810 (3)} & & \multicolumn{1}{c}{15 (2)} & \multicolumn{1}{c}{694 (2)} \\
     \bottomrule    
   \end{tabular}}
\footnotesize{$^1$ True Label, $^2$ Predicted Label}
\end{table*}
\renewcommand{\baselinestretch}{1.5} 
The results present in Table~\ref{tab:confusionmatrix} for Case 1, show that the samples of the normal group were all correctly classified. The anomalies had an average classification error of 25 samples in a total of 453 anomalies, confirming the good performance of the model. For Case 2, on average, 2 anomalies of 812 were classified incorrectly, and 2 normal samples of 24 were classified as anomalies. For Case 3, 694 of 709 anomalies were classified correctly and 82 of 107 normal samples were also  classified correctly. As in Case 1 and 2, the results show the good performance of the model.

Such performances in a real application, will allow not to intervene in the machine unnecessarily (which is also a big problem, considering the need to stop the production and high cost of some components that could be replaced without need). Moreover, the model was able to correctly identify most anomalies, including those at an early stage of fault, allowing the maintenance team to schedule the machine shutdown without directly interfering in the production process.

\begin{figure}[ht]
\begin{subfigure}{1\textwidth}
  \centering
  \includegraphics[width=.6\linewidth]{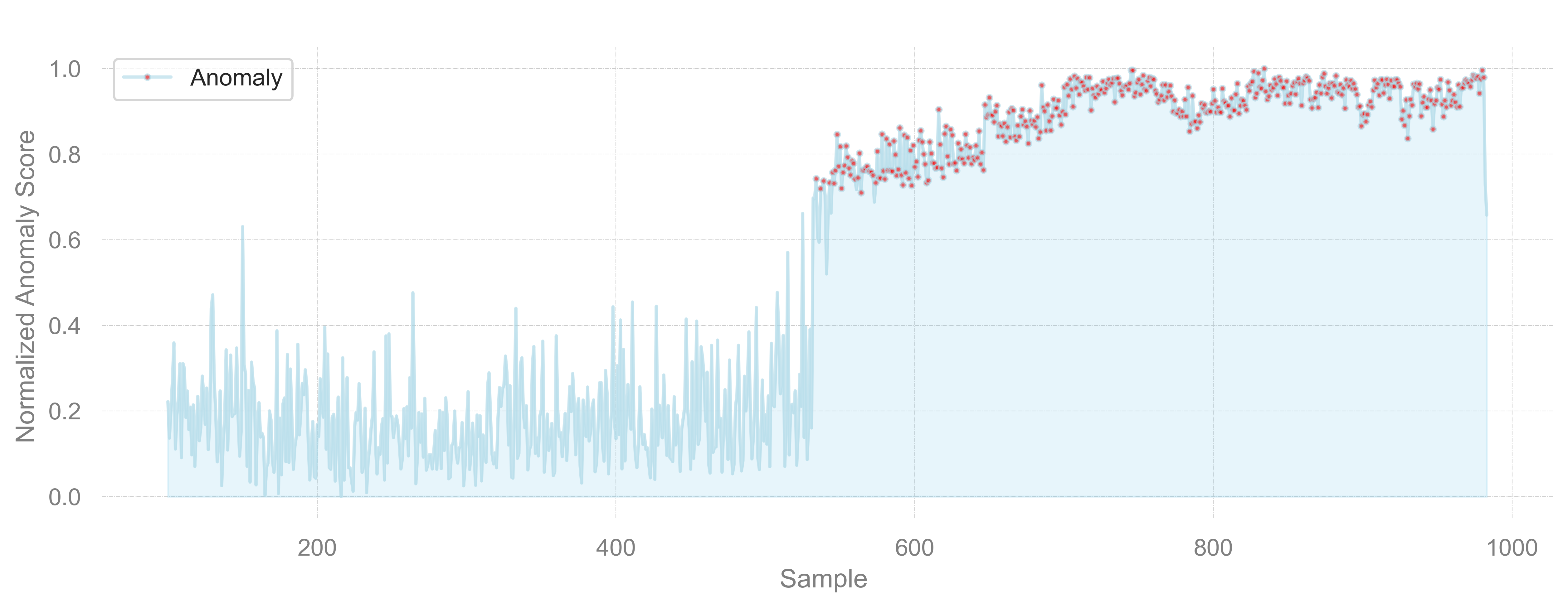}  
  \caption{Case 1 - Bearing Dataset.}
  \label{fig:sub-first}
\end{subfigure}
\newline
\begin{subfigure}{1\textwidth}
  \centering
  \includegraphics[width=.6\linewidth]{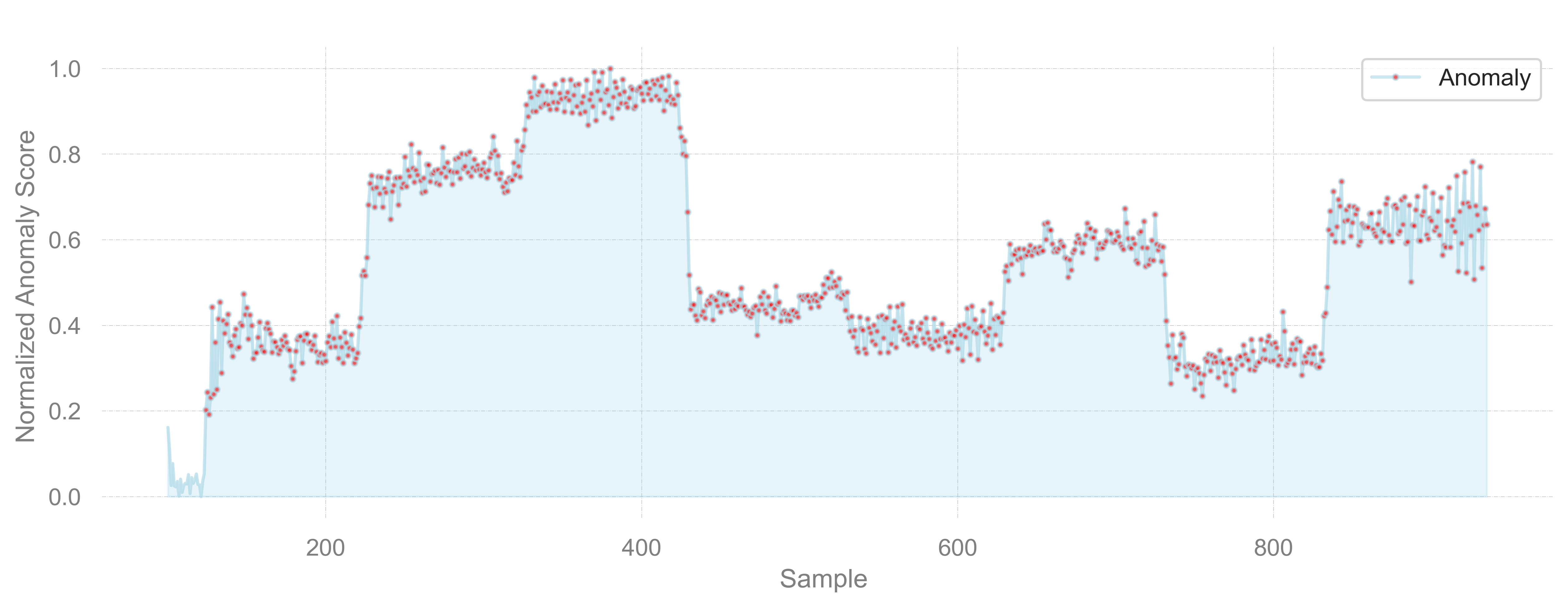}  
  \caption{Case 2 - Gearbox Dataset.}
  \label{fig:sub-sec}
\end{subfigure}
\newline
\begin{subfigure}{1\textwidth}
  \centering
  \includegraphics[width=.6\linewidth]{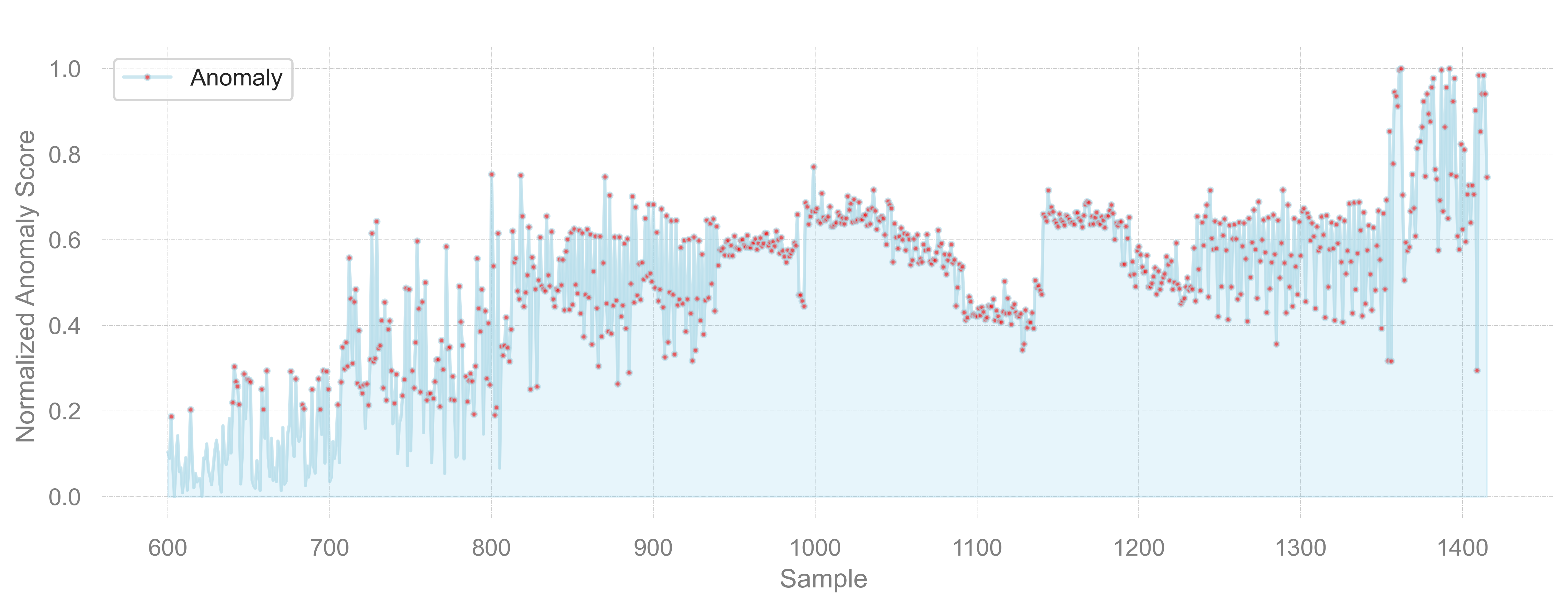}  
  \caption{Case 3 - Mechanical Fault Dataset.}
  \label{fig:sub-thr}
\end{subfigure}
\caption{Anomaly scores for Isolation Forest.}
\label{fig:anomalyscore}
\end{figure}

Keeping in mind that the essence of anomaly detection methods is unsupervised, that is, without defining even the threshold value (in addition to not having labelled data in training), the normalized anomaly scores are presented for the entire test, Fig. \ref{fig:anomalyscore}. The anomalies identified in the Fig. \ref{fig:anomalyscore} are presented based on the defined threshold. The x-axis values refer to the test samples only.
\renewcommand{\baselinestretch}{1.3} 
\renewcommand{\baselinestretch}{1.5} 
Fig. \ref{fig:sub-first} shows the evolution of the anomaly score with the development of the fault. It is possible to notice the gradual increase near the region identified as the beginning of an incipient defect, sample 531, as shown in Fig. \ref{fig:cascata}. Subsequently, there is an increase in the anomaly score in relation to the normal condition, indicating a permanent change in the behavior of the equipment, and consequently, a fault.

For Case 2 and 3, the anomaly score value does not show the evolution of the fault, since the analysis was performed in a static way. Analyzing Case 2, it is noted that the scores for samples considered anomalies are higher than the normal group (first 24 samples), enabling identification. The samples were grouped in sequence with respect to the equipment condition to allow comparison between the faults (healthy condition, missing tooth, root crack, spalling, chipping tip 1, chipping tip 2, chipping tip 3, chipping tip 4, chipping tip 5). It can be seen that similar faults with similar condition have similar anomaly scores. Thus, in a real monitoring situation, if the anomaly score changes suddenly, it can be concluded that a possible new fault is occurring or that the severity of the current fault has been accentuated.

For Case 3, due to the presence of different fault conditions in each group, it is not possible to distinguish the type of fault through visual analysis of the anomaly score. However, the variation between normal samples (first 107 samples) and those considered to be fault still differ, even if less than the other cases.

Despite the defined threshold value for comparison of the models, it is possible to notice the difference between normal and faulty samples, even for where the fault is incipient. This difference allows the user to correctly identify anomalies present in the monitoring. The importance of anomalies detected in the incipient period is emphasized, as it is a stage where it is not easily identified by visual analysis or variation of the time signal energy trend (often used as a metric to define maintenance alarm levels based on standards).

The difference in the anomaly score values also prove that the faults studied for rotating machinery behave as anomalies / outliers. In other words, they are samples that have values so different from other observations that they are capable of raising suspicions about the mechanism from which they were generated \cite{hawkinsbook}.  As the samples share this basic principle, it is concluded that the use of AI models based on anomaly detection allows to identify faults in rotating machinery in a satisfactory and unsupervised way.

\subsection{Fault Diagnosis: Unsupervised Classification / Root Cause Analysis}

The results obtained using the proposed methodology for the unsupervised classification are presented in the Table~\ref{tab:unsupervised}. The definition of the type of fault diagnosis to be used is performed during the extraction of features based on the premise of the features being related to different types of fault or not.

Case 1 and 2 present specific features related to a single type of fault / location, which allow the identification of the type of fault and the location, respectively. Therefore, the proposed Unsupervised Classification can be performed.
\renewcommand{\baselinestretch}{1.3} 
\begin{table*}[ht]
   \caption{}
   \caption*{Fault Diagnosis: Unsupervised Classification}
   \label{tab:unsupervised}
   \resizebox{\textwidth}{!}{\begin{tabular}{cccccccccccc}
     \toprule
     Case & kNN & MCD & LOF & CBLOF & OCSVM & FB & FastABOD & IF & HBOS & LODA & Ensemble\\
     \midrule
     \textbf{Case 1} & BSF & BPFO & BPFO & BSF & BSF & \textbf{BPFO} & BPFO & \textbf{BPFO} & \textbf{BPFO} & BPFO & BPFO\\
     Accuracy & 39.46 & 82.23 & 85.44 & 4.95 & 7.74 & \textbf{95.43} & 75.22 & \textbf{99.57} & \textbf{99.38} & 88.82 & 83.80\\
     Std  & (7.60) & (1.08) & (1.71) & (1.06) & (0.69) & (0.99) & (1.84) & (0.58) & (0.19) & (3.05) & (3.38)\\
      Time [s] & 2.6094 & \textbf{0.0968} & 0.2912 & \textbf{0.1913} & 0.2571 & 0.9428 & 4.3812 & 0.2890 & \textbf{0.1564} & 0.2190 & 9.7058\\
     \textbf{Case 2} & \textbf{1\textsuperscript{st}Stage} & 1\textsuperscript{st}Stage & 1\textsuperscript{st}Stage & \textbf{1\textsuperscript{st}Stage} & 1\textsuperscript{st}Stage & 1\textsuperscript{st}Stage & 1\textsuperscript{st}Stage & 1\textsuperscript{st}Stage & 1\textsuperscript{st}Stage & 1\textsuperscript{st}Stage & \textbf{1\textsuperscript{st}Stage}\\
     Accuracy & \textbf{96.47} & 58.13 & 84.90 & \textbf{93.42} & 69.18 & 91.09 & 88.91 & 86.12 & 86.84 & 89.83 & \textbf{96.72}\\
     Std  & (0.77) & (12.64) & (2.88) & (2.19) & (3.61) & (3.27) & (3.83) & (3.06) & (3.88) & (8.31) & (1.11)\\
       Time [s] & 2.3111 & \textbf{0.1774} & 0.3556 & \textbf{0.1998} & 0.2033 & 0.9097 & 7.2525 & 0.2538 & \textbf{0.1939} & 0.2644 & 12.6619\\
     \bottomrule
   \end{tabular}}
\end{table*}
\renewcommand{\baselinestretch}{1.5} 
For Case 1, it can be noted that IF, HBOS and FB models had better results. Using IF as an example, in 99.57\% of the samples analyzed, the specific feature BPFO was considered the most relevant, and consequently, correctly classifying the type of fault. Analyzing the results obtained in the previous stage of fault detection, IF and HBOS are good models for the methodology, since they showed good ability to detect and diagnose the fault. FB on the other hand, notwithstanding a good result in diagnosis, presented a low fault detection rate, which in this case would fail to identify some anomalies in the equipment. Some models such as CBLOF, kNN and OCSVM classified the fault as BSF instead of BPFO, being considered an error. The other models, despite having correctly classified the type of bearing fault, had a lower hit rate than those mentioned above, both in the fault detection stage and in the unsupervised classification.

For Case 2, the fault was classified in relation to the location in the gearbox. The most relevant features were associated according to their stage. In other words, using the Ensemble model as an example, in 96.72\% of the samples analyzed, the most relevant feature was related to fault in the first stage. The models with the highest hit rate were CBLOF, kNN and Ensemble. The models showed good results for both fault detection and diagnosis. It is worth mentioning that for the dataset under analysis, most models showed good results in detecting faults, possibly because they have well-characterized behaviors. IF and HBOS which presented good results for the fault detection in all cases, showed inferior performance, erroneously classifying approximately 15\% of the fault as present in the second stage. In general, the MCD that showed good results for fault detection, was not as effective in the fault diagnosis part.

For Case 3, the features may be related to more than one fault, therefore, it is not possible to perform the unsupervised classification directly. In this case, the general analysis, using the Root Cause Analysis procedure is applied, Table~\ref{tab:rca}.
\renewcommand{\baselinestretch}{1.3} 
\begin{table*}[ht]
   \caption{}
   \caption*{Fault Diagnosis: Root Cause Analysis results}
   \label{tab:rca}
   \resizebox{\textwidth}{!}{\begin{tabular}{cccccccccccc}
     \toprule
     Case & kNN & MCD & LOF & CBLOF & OCSVM & FB & FastABOD & IF & HBOS & LODA & Ensemble\\
     \midrule
     \textbf{Case 3.1$^1$} & 3xfr & 2xfr & 3xfr & 3xfr & 1xfr & 4xfr & 4xfr & 1xfr & 3xfr & 2xfr & 1xfr\\
      & 33.17 & 36.29 & 46.90 & 36.00 & 62.60 & 42.61 & 49.64 & 55.83 & 50.85 & 41.29 & 27.22\\
      & (0.00) & (5.10) & (0.00) & (6.07) & (0.00) & (4.47) & (0.00) & (6.49) & (0.00) & (14.92) & (2.54)\\
      Time [s] & 1.6272 & \textbf{0.1075} & 0.2681 & \textbf{0.1389} & 0.1981 & 0.8740 & 4.3299 & 0.3905 & \textbf{0.1301} & 0.2256 & 8.3899\\
     \textbf{Case 3.2$^2$} & 3xfr & 3xfr & 3xfr & 2xfr & 3xfr & 2xfr & 3xfr & 2xfr & 3xfr & 3xfr & 3xfr\\
      & 44.09 & 35.28 & 49.74 & 49.84 & 68.25 & 34.13 & 39.80 & 45.24 & 61.08 & 53.00 & 43.13\\
      & (0.00) &  (4.93) & (0.00) & (3.27) & (0.00) & (10.21) & (0.00) & (7.94) & (0.00) & (11.11) & (3.14)\\
      Time [s] & 1.3873 & \textbf{0.1207} & 0.2199 & \textbf{0.1373} & 0.2021 & 0.8537 & 4.3308 & 0.3436 & \textbf{0.1136} & 0.1607 & 8.0897\\
     \textbf{Case 3.3$^3$} & 2xfr & 1xfr & 4xfr & 2xfr & 1xfr & 2xfr & 2xfr & 2xfr & 2xfr & 1xfr & 2xfr\\
      & 56.52 & 47.36 & 39.78 & 42.55 & 66.67 & 51.26 & 60.86 & 45.53 & 47.82 & 49.70 & 25.73\\
      & (0.00) & (0.00) & (0.00) & (13.83) & (0.00) & (27.57) & (0.00) & (19.51) & (0.00) & (19.21) & (7.02)\\
      Time [s] & 1.3114 & \textbf{0.1210} & 0.1941 & \textbf{0.1212} & 0.1799 & 0.8174 & 3.6984 & 0.2489 & \textbf{0.1136} & 0.1557 & 7.1616\\
    \textbf{Case 3.4$^4$} & 2xfr & 3xfr & 1xfr & 2xfr & 2xfr & 1xfr & 4xfr & 2xfr & 4xfr & 4xfr & 1xfr\\
      & 66.66 & 64.28 & 46.66 & 38.53 & 73.33 & 41.20 & 66.66 & 42.93 & 86.66 & 43.94 & 50.00\\
      & (0.00) & (0.00) & (0.00) & (10.77) & (0.00) & (15.18) & (0.00) & (28.14) & (0.00) & (20.13) & (10.34)\\
      Time [s] & 1.5150 & \textbf{0.1303} & 0.2225 & \textbf{0.1144} & 0.1740 & 0.8191 & 3.8296 & 0.2812 & \textbf{0.1198} & 0.1586 & 7.5644\\
     \bottomrule
   \end{tabular}}
\footnotesize{$^1$ Unbalance, $^2$ Misalignment, $^3$ Mechanical Looseness, $^4$ Combined Faults}
\end{table*}
\renewcommand{\baselinestretch}{1.5} 

For better visualization, the results are presented based on the most relevant feature obtained by the methodology. A sub-division for each type of fault was carried out in order to provide more details on the method. An example of the complete results is presented for the IF and Case 3.1, Table~\ref{tab:fullranking}.

The unbalance fault is presented in Case 3.1 and the results are shown in Table~\ref{tab:rca} and Table~\ref{tab:fullranking}. Due to the unbalance behavior predominantly manifesting in 1xfr, it is expected that this features will show greater relevance for the analysis, as presented in the IF and OCSVM models. On the other hand, as the features are directly or indirectly related to more than one fault, the model can use the relationship with another feature, instead of what is expected. For example: it is known that unbalance manifests itself in 1xfr, however, if the energy in 2xfr is greater than 1xfr, possibly the sample presents a misalignment (excluding other fault possibilities just for example). Thus, assuming an unbalanced sample, the model can use 2xfr, as a basis to know if it is less or greater than 1xfr and thus 2xfr becomes the most relevant feature, even if the fault is an unbalance. In addition to the aforementioned justification, the type of fault introduced was considered to label the samples. Thus, in some cases the fault behavior was not evident in the signal, which justifies the model to identify other features as more relevant. For example: for a small unbalance, the acquired signal is considered to be unbalanced, even if it does not significantly increase the amplitude in 1xfr.

Table~\ref{tab:fullranking} shows that in 55.83 \% of the samples, 1xfr was classified as the most relevant feature. Subsequently, the features 2x and 3xfr are the most important. Such features are related to the way of identifying an unbalance in a vibration signal, and therefore they can be used by the specialist to analyze the root cause of the fault. It is also noted that the 4xfr feature in most cases was classified as less relevant, since the feature (for the case under study) is not so important for identifying or distinguishing this fault.
\renewcommand{\baselinestretch}{1.3} 
\begin{table*}[ht]
   \caption{}
   \caption*{Fault Diagnosis: Root Cause Analysis full ranking}
   \label{tab:fullranking}
    \begin{tabu} to \textwidth {X[c]X[c]X[c]X[c]X[c]}
     \toprule
     Feature/Position & 1$^{st}$ & 2$^{nd}$ & 3$^{rd}$ & 4$^{th}$ \\
     \midrule
     1xfr & 55.83 (6.50) & 20.16 (1.86) & 16.00 (4.92) & 8.01 (3.66)\\
     2xfr & 15.97 (3.20) & 45.89 (10.14) & 27.83 (9.09) & 10.31 (2.82)\\
     3xfr & 19.50 (5.40) & 26.87 (9.57) & 35.88 (9.54) & 17.75 (4.78)\\
     4xfr & 8.70 (2.32) & 7.08 (1.86) & 20.29 (4.07) & 63.93 (5.08)\\
     \bottomrule
   \end{tabu}
\end{table*}
\renewcommand{\baselinestretch}{1.5} 
In Case 3.2 the misalignment is presented. Usually this type of fault is identified by the analysis of 2xfr and 3xfr. All models presented, as the most important feature, the same one used by the human specialist.

The mechanical looseness, Case 3.3, as well as the combination of faults, Case 3.4, can present energy in all extracted features. Thus, the variation of the features selected by each model is acceptable, since all features are relevant. It is noteworthy that the variation between the models is due to the different approaches present in each algorithm.

The importance of root cause analysis is to eliminate features that are not relevant to the analysis, helping the specialist to identify the problem. Thus, in an application where different features and faults are present, the methodology provides a better direction to the specialist about the current fault.

The standard deviation presented in some analysis in Case 3, can be justified by the random selection of samples at each iteration. As mentioned earlier, in addition to the different possibilities of the model in relating the features to the faults, the samples can also present different behaviors, even within the same type of faults, resulting in different selected features. As some models have stochastic behavior, they are more sensitive to variation. Note that for Cases 1 and 2, where there are no major variations in relation to the type of fault, the models have low standard deviations. 

The models with the lowest computational costs for the fault diagnosis methodology were: CBLOF, HBOS and MCD, with HBOS being one of the fastest also for fault detection.

Through the explainability of the artificial intelligence models used, it can be concluded that the proposed methodology is able to assist the specialist in identifying the root cause of the problem or even to classify the type of fault present in the equipment in an unsupervised way.

\subsection{XAI: SHAP and Local-DIFFI}

As presented in the methodology, the unsupervised classification/root cause analysis is performed through the ranking of importance of the specific features obtained by the model's explainability. To study the possibility of the methodology in working with different explainable models and the feasibility of implementing a computationally faster model, in Table~\ref{tab:xairanking} is shown a comparison for the complete relevance rankings obtained by SHAP and Local-DIFFI. As the main goal is to compare the two methods, the values for Case 3 were calculated for all faults. From Table~\ref{tab:xairanking}, the time taken to perform the explainability was higher using SHAP than Local-DIFFI. As a model-specific, Local-DIFFI presents a superior performance of approximately 6.5-8.0x in relation to SHAP, being extremely relevant in applications where the execution time is essential. 
\renewcommand{\baselinestretch}{1.3} 
\begin{table*}[ht]
   \caption{}
   \caption*{XAI: SHAP vs. Local-DIFFI}
   \label{tab:xairanking}
    \begin{tabu} to \textwidth {X[c]X[c]X[c]X[c]}
     \toprule
     Metric/Case & Case 1 & Case 2 & Case 3 \\
     \midrule
     Kendall-Tau Distance & 0.348 & 0.127 & 0.455\\
     SHAP: Time [s] & 0.2890 & 0.2538 & 0.3012\\
     Local-DIFFI: Time [s] & 0.0361 & 0.0365 & 0.0453\\
     \bottomrule
   \end{tabu}
\end{table*}
\renewcommand{\baselinestretch}{1.5} 

\begin{figure}[ht]
  \subfloat[Case 1 - SHAP]{
	\begin{minipage}[c][0.9\width]{
	   0.3\textwidth}
	   \centering
    \includegraphics[width=1\textwidth]{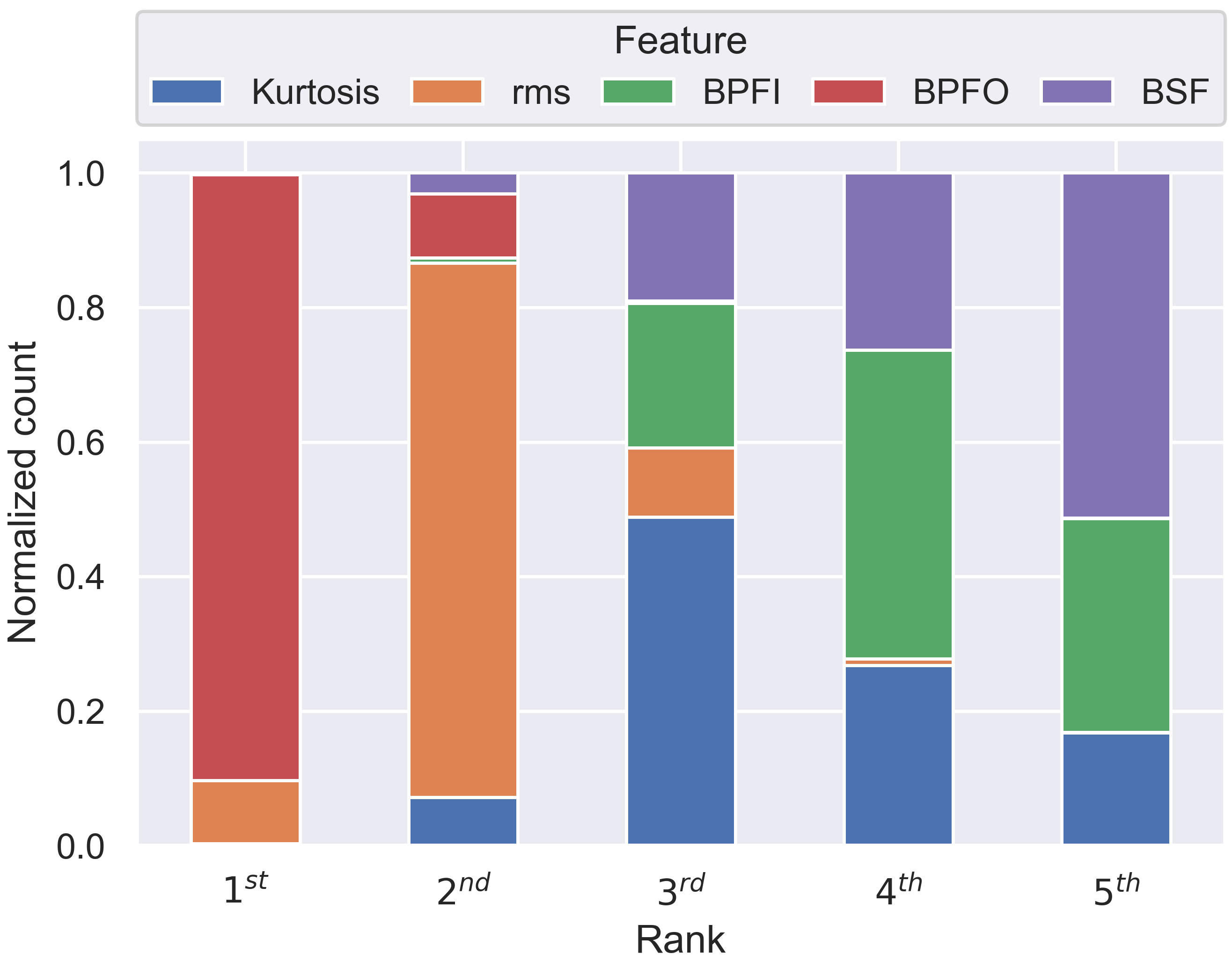}
	\end{minipage}}
 \hfill 	
  \subfloat[Case 2 - SHAP]{
	\begin{minipage}[c][0.9\width]{
	   0.3\textwidth}
	   \centering
	   \includegraphics[width=1\textwidth]{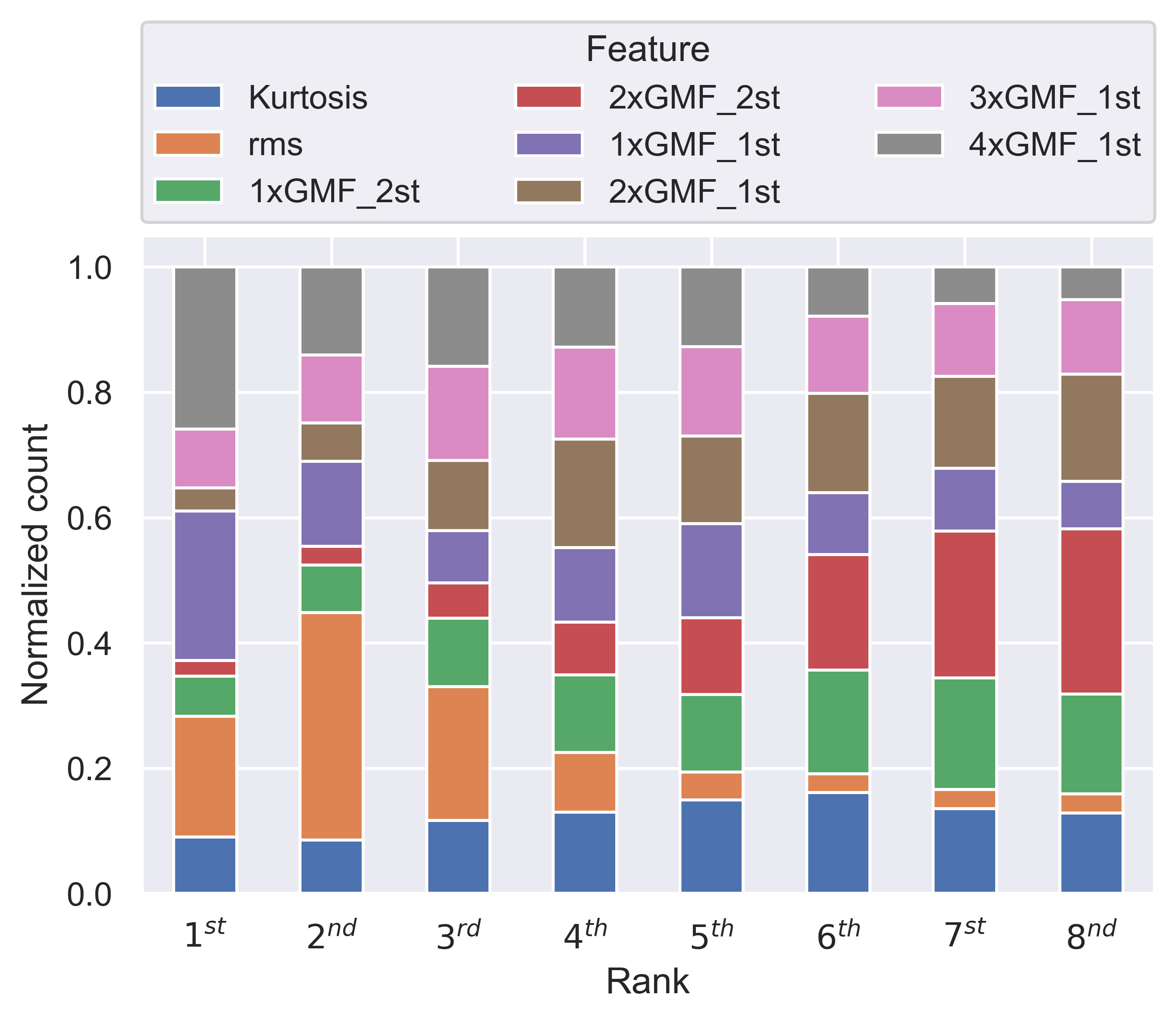}
	\end{minipage}}
 \hfill	
  \subfloat[Case 3 - SHAP]{
	\begin{minipage}[c][0.9\width]{
	   0.3\textwidth}
	   \centering
	   \includegraphics[width=1\textwidth]{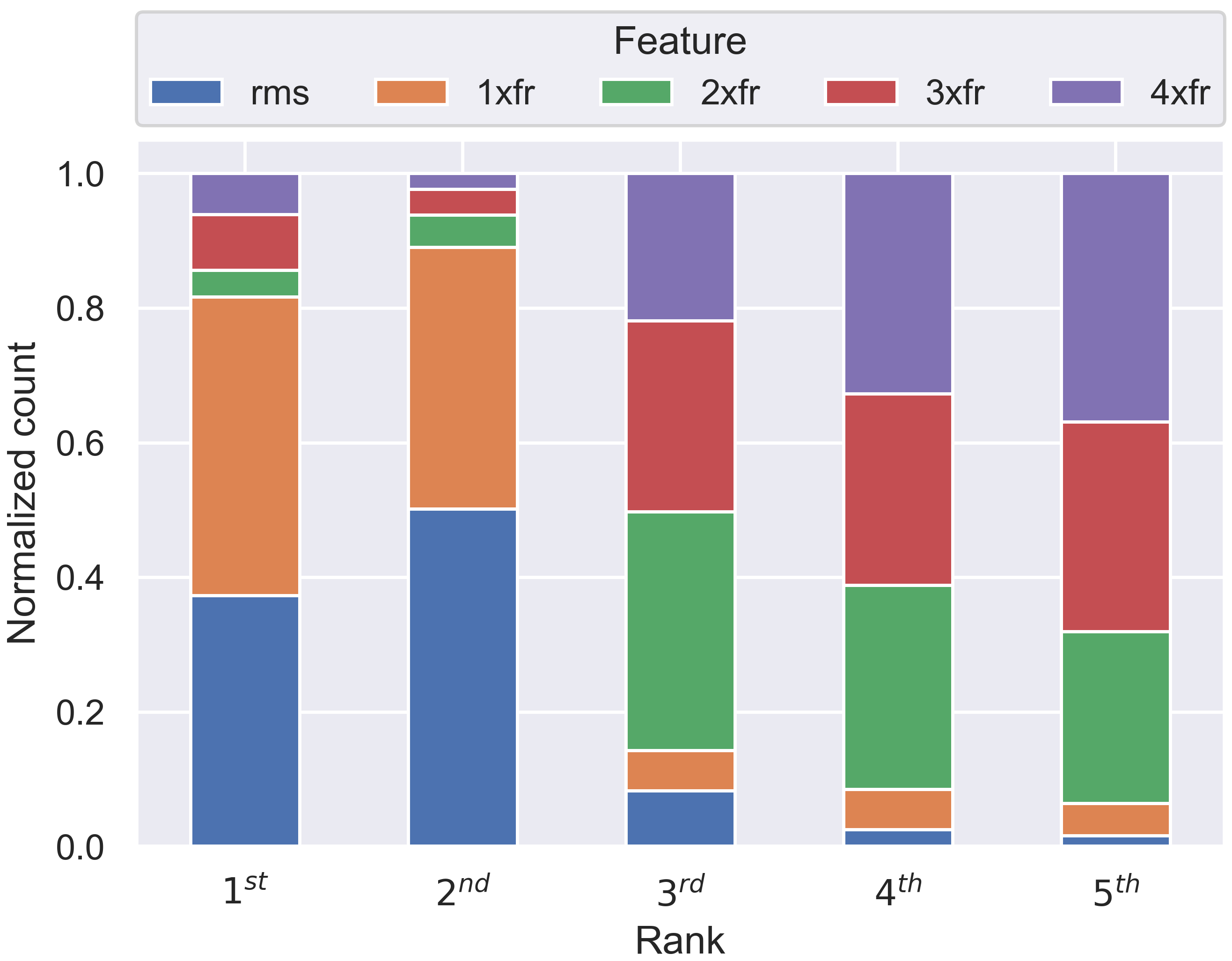}
	\end{minipage}}\\
    
  \subfloat[Case 1 - Local-DIFFI]{
	\begin{minipage}[c][0.9\width]{
	   0.3\textwidth}
	   \centering
	   \includegraphics[width=1\textwidth]{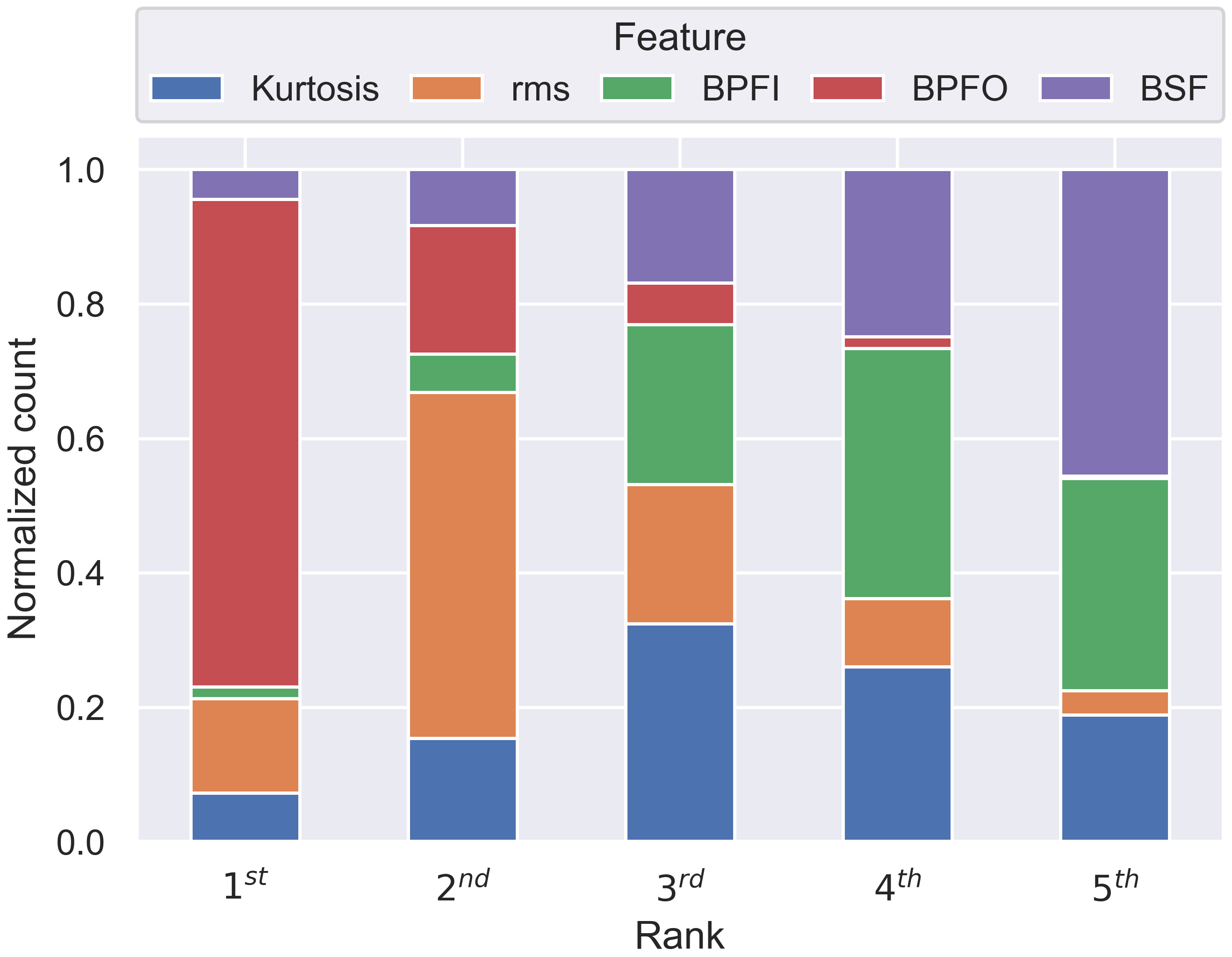}
	\end{minipage}}
 \hfill 	
  \subfloat[Case 2 - Local-DIFFI]{
	\begin{minipage}[c][0.9\width]{
	   0.3\textwidth}
	   \centering
	   \includegraphics[width=1\textwidth]{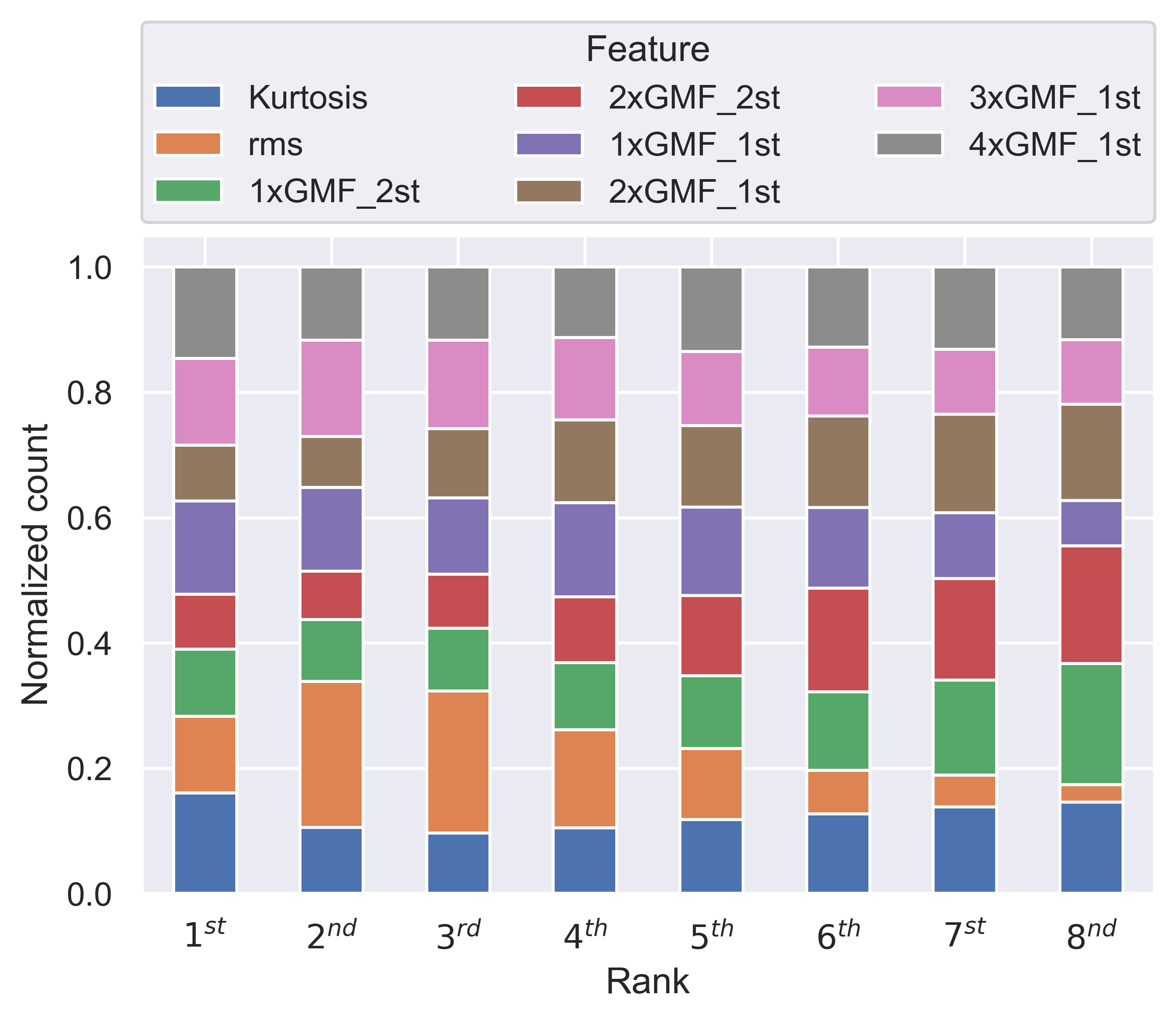}
	\end{minipage}}
 \hfill	
  \subfloat[Case 3 - Local-DIFFI]{
	\begin{minipage}[c][0.9\width]{
	   0.3\textwidth}
	   \centering
	   \includegraphics[width=1\textwidth]{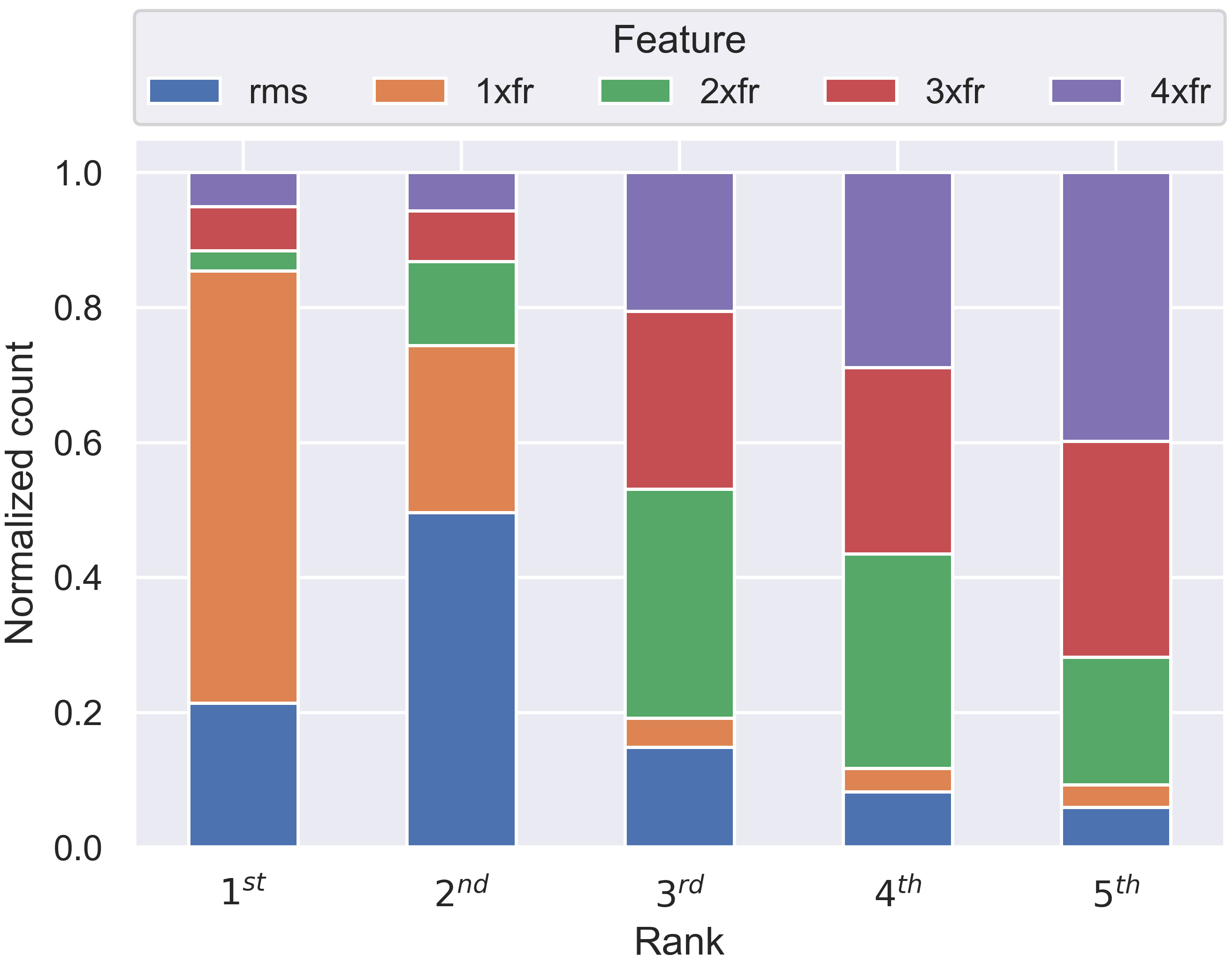}
	\end{minipage}}
    \caption{SHAP and Local-DIFFI feature importance ranking.}
    \label{fig:rankingdiffi}
\end{figure}

The comparison made through Kendall-Tau distance shows that the models have similarities in the rankings of relevance, visually presented in Fig. \ref{fig:rankingdiffi}. Since the main objective is to compare the two models, all the features used by the models for fault detection are considered, without excluding the general features proposed in the application of the fault diagnosis part.
\renewcommand{\baselinestretch}{1.3} 
\renewcommand{\baselinestretch}{1.5} 
It can be seen in Fig. \ref{fig:rankingdiffi}, for Case 1, that the most relevant feature for both models is precisely BPFO, allowing the unsupervised classification to achieve good results. For Local-DIFFI, some samples presented BPFI and BSF as the most important specific feature, leading the methodology to misclassify the type of fault. Case 2 presented the lowest relationship between the rankings. Among the most relevant features, SHAP showed less occurrence of the features related to the second stage than Local-DIFFI, leading the model to make fewer errors during the application of the proposed methodology. Despite the minor similarity, the main feature (1xGMF\_1st) was also the same in both models. For Case 3, the most relevant feature for both models is 1xfr with a good similarity for positions 3, 4 and 5. Thus, through the analysis of the Kendall-Tau distance, it is possible to verify that the rankings show similar behaviors. Because it is a model-specific method, Local-DIFFI is subject to noise due to the stochasticity of the IF, which can reduce its result. Finally, the choice of the explainability model to be used is based on a trade-off between response time and precision. 

\section{Conclusions}

This paper proposes a new approach for fault detection and diagnosis in rotating machinery. A three-stage scheme is adopted 1) Feature extraction; 2) Fault detection: Anomaly Detection; 3) Fault diagnosis: Unsupervised classification / Root cause analysis. The vibration features in the time and frequency domains were extracted based on human knowledge already available. In the fault detection, the presence of fault was verified in an unsupervised manner based on anomaly detection algorithms. Finally, in fault diagnosis, through the feature importance ranking obtained by the model's explainability, the fault diagnosis was performed, being: unsupervised classification or root cause analysis.

The results show that the proposed methodology allows the unsupervised fault detection in rotating machinery. And, in addition to providing explainability about the models used, the methodology provides relevant information for root cause analysis, or even unsupervised fault classification. 

Different state-of-the-art ML algorithms in anomaly detection were studied showing the possibility to change models according to the dataset. The new approach can be applied to different types of faults just by modifying the extracted features associated with a potential fault as shown for the 3 datasets studied. Since the approach does not require previously labeled data, and only knowledge currently available on fault detection through vibration analysis, the methodology has many possible industrial applications. Future work will focus on domain adaptation and transfer learning associated with methods for model interpretability to improve the applicability of the proposed approach in different industrial scenarios.

\section*{Acknowledgement}
\addcontentsline{toc}{section}{Acknowledgement}

The authors gratefully acknowledge the Brazilian research funding agencies CNPq (National Council for Scientific and Technological Development) and CAPES (Federal Agency for the Support and Improvement of Higher Education) for their financial support of this work.

\addcontentsline{toc}{section}{References}
\bibliography{mybibfile}

\end{document}